\documentclass[10pt,journal,compsoc]{IEEEtran}

\usepackage{cite}
\usepackage{graphicx}
\usepackage{algorithmic}
\usepackage{algorithm}
\usepackage{amssymb, amsmath}
\usepackage{booktabs}
\usepackage{color}
\usepackage{threeparttable}
\usepackage{array}
\usepackage{amsfonts}
\usepackage{subfigure}
\usepackage{multirow}
\usepackage{threeparttable}
\usepackage{array}
\usepackage{cite}
\usepackage{ulem}
\usepackage{etoolbox}
\usepackage{ragged2e}
\usepackage{float}
\usepackage{enumerate}
\usepackage{tikz}
\usepackage{comment}
\usepackage{makecell}
\usepackage{graphicx}
\usepackage[pagebackref,breaklinks,colorlinks]{hyperref}

\renewcommand{\raggedright}{\leftskip=0pt \rightskip=0pt plus 0cm}

\makeatletter
\newcommand{\printfnsymbol}[1]{%
  \textsuperscript{\@fnsymbol{#1}}%
}
\makeatother

\usepackage{multirow}

\usepackage{enumitem}
\usepackage{multicol}

\graphicspath{{image/}}

\usepackage{cases}
\usepackage{bm}
\usepackage{cases}
\usepackage{multirow}
\usepackage{color} 
\usepackage{enumerate}
\usepackage{etoolbox}
\usepackage{mathrsfs}

\newcommand{\MyMapTemplatePrefix}[4]{\expandafter#1\csname#3#4\endcsname{#2{#4}}}
\newcommand{\MyMapTemplatePrefixNew}[5]{\expandafter#1\csname#4#5\endcsname{#2{#3{#5}}}}
\forcsvlist{\MyMapTemplatePrefix {\def} {\mathbf} {}} {A,B,C,D,E,F,G,H,I,J,K,L,M,N,O,P,Q,R,S,T,U,V,W,X,Y,Z}
\forcsvlist{\MyMapTemplatePrefix {\def} {\mathbf} {}} {a,b,c,d,e,f,g,h,i,j,k,l,m,n,o,p,q,r,s,t,u,v,w,x,y,z,1,0}
\forcsvlist{\MyMapTemplatePrefix {\def} {\widetilde} {wt}} {A,B,C,D,E,F,G,H,I,J,K,L,M,N,O,P,Q,R,S,T,U,V,W,X,Y,Z}
\forcsvlist{\MyMapTemplatePrefix {\def} {\widetilde} {wt}} {a,b,c,d,e,f,g,h,i,j,k,l,m,n,o,p,q,r,s,t,u,v,w,x,y,z} 
\forcsvlist{\MyMapTemplatePrefixNew {\def} {\widetilde}{\mathbf} {tb}} {A,B,C,D,E,F,G,H,I,J,K,L,M,N,O,P,Q,R,S,T,U,V,W,X,Y,Z}
\forcsvlist{\MyMapTemplatePrefixNew {\def} {\widetilde}{\mathbf} {tb}} {a,b,c,d,e,f,g,h,i,j,k,l,m,n,o,p,q,r,s,t,u,v,w,x,y,z}
\forcsvlist{\MyMapTemplatePrefix {\def} {\widehat} {wh}} {A,B,C,D,E,F,G,H,I,J,K,L,M,N,O,P,Q,R,S,T,U,V,W,X,Y,Z}
\forcsvlist{\MyMapTemplatePrefix {\def} {\widehat} {wh}} {a,b,c,d,e,f,g,h,i,j,k,l,m,n,o,p,q,r,s,t,u,v,w,x,y,z}
\forcsvlist{\MyMapTemplatePrefixNew {\def} {\widehat}{\mathbf} {hb}} {A,B,C,D,E,F,G,H,I,J,K,L,M,N,O,P,Q,R,S,T,U,V,W,X,Y,Z}
\forcsvlist{\MyMapTemplatePrefixNew {\def} {\widehat}{\mathbf} {hb}} {a,b,c,d,e,f,g,h,i,j,k,l,m,n,o,p,q,r,s,t,u,v,w,x,y,z}
\forcsvlist{\MyMapTemplatePrefixNew {\def} {\overline}{\mathbf} {lb}} {A,B,C,D,E,F,G,H,I,J,K,L,M,N,O,P,Q,R,S,T,U,V,W,X,Y,Z}
\forcsvlist{\MyMapTemplatePrefixNew {\def} {\overline}{\mathbf} {lb}} {a,b,c,d,e,f,g,h,i,j,k,l,m,n,o,p,q,r,s,t,u,v,w,x,y,z}
\forcsvlist{\MyMapTemplatePrefix {\def} {\mathcal}{mc}} {A,B,C,D,E,F,G,H,I,J,K,L,M,N,O,P,Q,R,S,T,U,V,W,X,Y,Z}
\forcsvlist{\MyMapTemplatePrefix {\def} {\mathbb} {mb}} {A,B,C,D,E,F,G,H,I,J,K,L,M,N,O,P,Q,R,S,T,U,V,W,X,Y,Z}
\forcsvlist{\MyMapTemplatePrefix {\DeclareMathOperator} {} {} } {tr,diag,sgn}

\def\sha{\text{com}}
\def\prt{\text{prt}}

\allowdisplaybreaks[4]

\begin{document}

\title{Decoupled Hierarchical Distillation for Multimodal Emotion Recognition}

\author{Yong Li,
        Yuanzhi Wang,
        Yi Ding,
        Shiqing Zhang,
        Ke Lu,
        Cuntai Guan~\IEEEmembership{Fellow,~IEEE}

\IEEEcompsocitemizethanks{
\IEEEcompsocthanksitem Corresponding author: Cuntai Guan, Ke Lu, Shiqing Zhang \protect
\IEEEcompsocthanksitem Yong Li is with the School of Computer Science and Engineering, and the Key Laboratory of New Generation Artificial Intelligence Technology and Its Interdisciplinary Applications, Southeast University, Nanjing 210096, China. Email: mysee1989@gmail.com. 

Yuanzhi Wang is with the School of Computer Science and Engineering, Nanjing University of Science and Technology, Nanjing, 210094, China. E-mail: yuanzhiwang@njust.edu.cn. \protect

Shiqing Zhang is with the Institute of Intelligent Information Processing, Taizhou University, Taizhou 318000, Zhejiang, China. E-mail: tzczsq@163.com. \protect

Ke Lu is with the School of Engineering Science, University of Chinese Academy of Sciences, Beijing 100049, China, and also with the Peng Cheng Laboratory, Shenzhen, Guangdong 518055, China. E-mail: luk@ucas.ac.cn. \protect

\IEEEcompsocthanksitem Yi Ding and Cuntai Guan are with the School of Computer Science and Engineering, Nanyang Technological University, 50 Nanyang Avenue, Singapore, 639798. E-mail: {(ding.yi, ctguan)}@ntu.edu.sg. \protect \\ 
}
}

\markboth{Journal of \LaTeX\ Class Files,~Vol.~14, No.~8, August~2015}%
{Shell \MakeLowercase{\textit{et al.}}: Bare Demo of IEEEtran.cls for Computer Society Journals}

\IEEEtitleabstractindextext{%
\begin{abstract}
\raggedright{
Human multimodal emotion recognition (MER) seeks to infer human emotions by integrating information from language, visual, and acoustic modalities. Although existing MER approaches have achieved promising results, they still struggle with inherent multimodal heterogeneities and varying contributions from different modalities. To address these challenges, we propose a novel framework, Decoupled Hierarchical Multimodal Distillation (DHMD). DHMD decouples each modality's features into modality-irrelevant (homogeneous) and modality-exclusive (heterogeneous) components using a self-regression mechanism. The framework employs a two-stage knowledge distillation (KD) strategy: (1) coarse-grained KD via a Graph Distillation Unit (GD-Unit) in each decoupled feature space, where a dynamic graph facilitates adaptive distillation among modalities, and (2) fine-grained KD through a cross-modal dictionary matching mechanism, which aligns semantic granularities across modalities to produce more discriminative MER representations. This hierarchical distillation approach enables flexible knowledge transfer and effectively improves cross-modal feature alignment. 
Experimental results demonstrate that DHMD consistently outperforms state-of-the-art MER methods, achieving 1.3\%/2.4\% (ACC$_7$), 1.3\%/1.9\% (ACC$_2$) and 1.9\%/1.8\% (F1) relative improvement on CMU-MOSI/CMU-MOSEI dataset, respectively. Meanwhile, visualization results reveal that both the graph edges and dictionary activations in DHMD exhibit meaningful distribution patterns across modality-irrelevant/-exclusive feature spaces.
}
\end{abstract}

}

\maketitle

\IEEEdisplaynontitleabstractindextext

\IEEEpeerreviewmaketitle

\IEEEraisesectionheading{\section{Introduction}\label{sec:introduction}}


\IEEEPARstart{T}{he} integration of emotional intelligence into artificial intelligence (AI) systems represents a pivotal advancement towards the realization of artificial general intelligence~\cite{yonck2020heart}. Presently,  the field of artificial emotional intelligence  is in its early stages and contemporary multimodal emotion recognition (MER) methodologies focus on analyzing emotional states from video data, capturing time-series information across multiple modalities, including language, acoustic signals, and visual inputs~\cite{PMR,MICA}. This multimodal approach enhances our understanding of human behaviors and intentions from an integrative perspective. In essence, the capability to comprehend and express emotions through various modalities will substantially enhance AI's interactions with humans and the environment. This capability underpins numerous applications, such as intelligent tutoring
\newpage
\noindent
systems~\cite{petrovica2017emotion}, autonomous driving~\cite{kraus2009cognition}, and robotics~\cite{liu2017facial}.

\begin{figure}[htb]
	\includegraphics[width=1.0\linewidth]{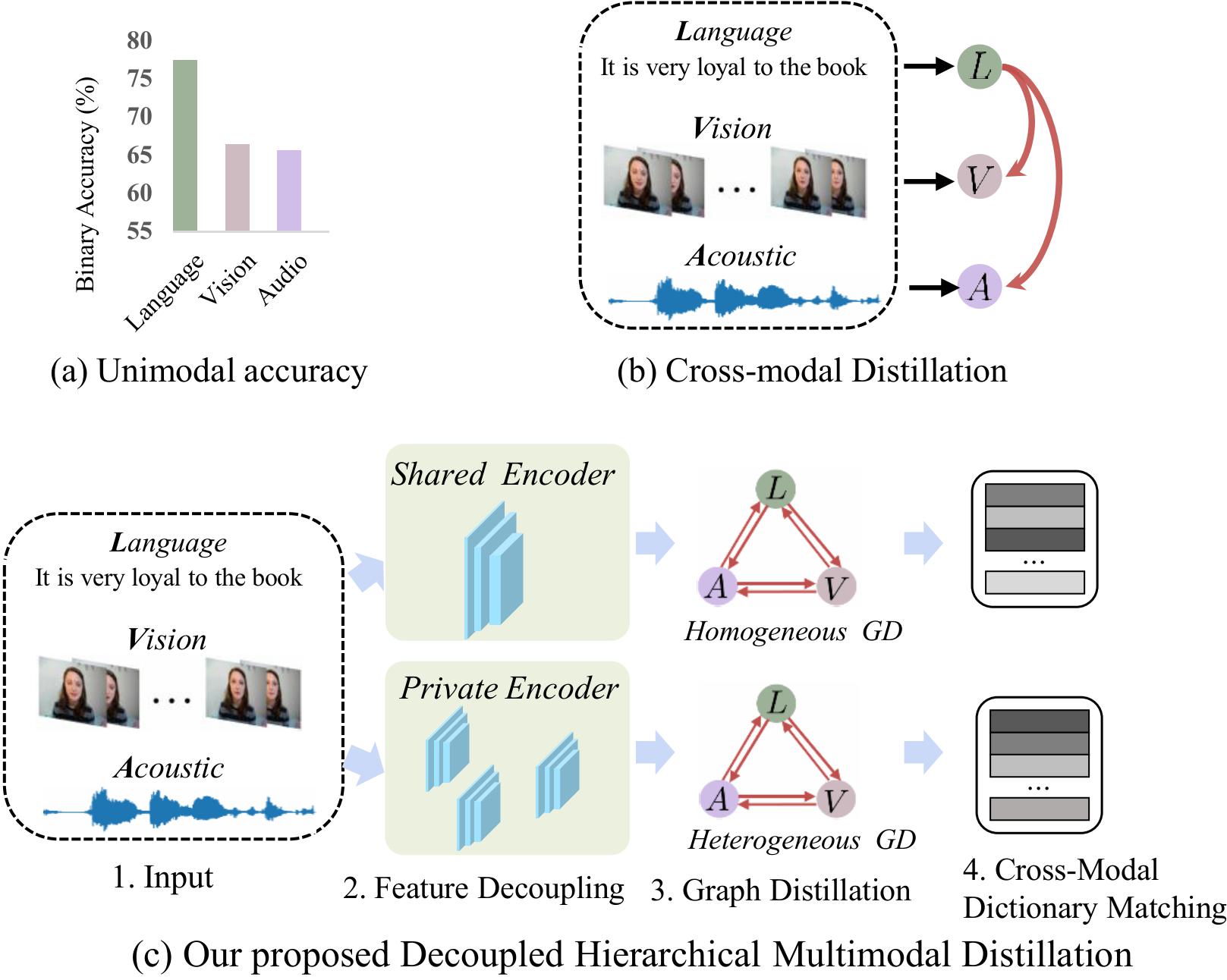}
	\caption{
		(a) illustrates the significant emotion recognition discrepancies using unimodality, adapted from Mult~\cite{MulT}. (b) shows the conventional cross-modal distillation. (c) shows our proposed DHMD method. DHMD implements a two-stage knowledge distillation (KD) approach comprising coarse- and fine-grained KD. Coarse-grained KD utilizes homogeneous and heterogeneous graph distillation units (GD) to reduce the complexity of cross-modal KD, enhancing specificity and efficiency. Fine-grained KD incorporates a shared dictionary as a unified semantic space for cross-modal alignment.
	}
	\label{fig:main_idea}
\end{figure}

In MER, different modalities within the same video segment often complement each other, offering additional cues for semantic and emotional clarification. The crux of MER lies in multimodal representation learning and fusion, where the objective is to encode and integrate representations from multiple modalities to discern the underlying emotions in raw data. Despite the progress achieved by prominent MER methods~\cite{TFN,MulT,MISA}, intrinsic heterogeneities among various modalities continue to pose significant challenges, complicating robust multimodal representation learning. Each modality, including image, language, and acoustic, conveys semantic information in distinct ways. Typically, the language modality comprises limited transcribed texts and embodies more abstract semantics compared to nonverbal behaviors~\cite{MulT}. As depicted in Fig.\ref{fig:main_idea} (a), language plays the most important role in MER and the intrinsic heterogeneities result in significant performance discrepancies among different modalities~\cite{MulT, pham2019found, TCSP}.

One straightforward approach to mitigating the pronounced modality heterogeneities in MER is to distill reliable and generalizable knowledge from stronger modalities to weaker ones~\cite{Crossmodaldistillation}. This concept, illustrated in Fig.~\ref{fig:main_idea} (b), leverages the strengths of each modality to enhance the overall model performance. However, manually assigning the direction or weights for distillation is cumbersome due to the multitude of possible combinations. Instead, models should autonomously adapt the distillation process based on different examples. For instance, many emotions are more easily recognized through language, while others are better captured visually. Furthermore, significant feature distribution mismatches across modalities can render direct cross-modal distillation sub-optimal.

As highlighted in~\cite{huang2021seeing, MulT}, cross-modal heterogeneity is intrinsic and induces the difficulties in multimodal representation learning, e.g., visual representation at pixel-level is much more diverse and dense than language embedding. Besides, the lack of explicit supervision for pixel-level language adds the difficulty to align different modalities.  This necessitates the development of sophisticated techniques that allow for automatic, example-specific distillation, aiming to improve the robustness and accuracy of the MER system. While cross-modal distillation holds promise for addressing modality heterogeneities, it requires adaptive mechanisms to account for the varying strengths of different modalities as well as the intrinsic mismatches in their feature distributions~\cite{zhang2021matching, nguyen2021knowledge}.

In this manuscript, we propose a Decoupled Hierarchical Knowledge Distillation (DHMD) method to mitigate the intrinsic cross-modal heterogeneity that impedes multimodal representation learning in MER, as depicted in Fig.~\ref{fig:main_idea} (c). In DHMD, features from each modality are decoupled into modality-irrelevant and modality-exclusive spaces via a shared encoder and private encoders, respectively. To achieve effective feature decoupling, we employ a self-regression mechanism that predicts the decoupled modality features and regresses them  self-supervisedly. To further reinforce feature decoupling, we introduce a margin loss that regularizes the relational proximity of representations across modalities and emotion categories. With the decoupled feature spaces, we are capable of conducting decoupled knowledge distillation (KD) in a more specialized and effective manner and reducing the complexity of assimilating knowledge from heterogeneous modalities. 

In DHMD, we introduce a two-stage KD framework within the decoupled multimodal feature spaces to establish cross-modal alignment in a hierarchical manner. In the first KD stage, a novel Graph Distillation Unit (GD-Unit) is employed in each feature space, facilitating coarse-grained cross-modal KD with increased specialization and efficacy. Each GD-Unit comprises a dynamic graph, where vertices represent different modalities and edges signify dynamic KD pathways. GD-Unit enables a flexible knowledge transfer mechanism, wherein distillation weights are automatically learned, supporting diverse cross-modal knowledge transfer patterns. As the distribution gap among modality-irrelevant (homogeneous) features is sufficiently minimized, GD can effectively capture inter-modality semantic correlations. For modality-exclusive (heterogeneous) features, a multimodal transformer is utilized to establish semantic alignment and bridge the distribution gap. 
For clarity, the graph distillation processes in the decoupled multimodal features are termed homogeneous graph knowledge distillation (HoGD) and heterogeneous graph knowledge distillation (HeGD). 

In the second stage, a cross-modal Dictionary Matching (DM) mechanism is proposed for fine-grained multimodal KD. Previous studies~\cite{huang2021seeing, duan2022multi} have highlighted that information in different modalities naturally exhibits varying levels of granularity\textemdash emotion cues, for instance, may be present in a few discrete words in the language modality or specific key facial frames in the visual modality. Drawing on this observation, we hypothesize that unifying the semantic granularities across various input modalities can yield more discriminative MER representations. To achieve this, a shared dictionary is incorporated, serving as a common semantic space to align different modalities within each of the decoupled feature spaces. Through cross-modal DM, modalities corresponding to the same emotion category are optimized to be represented using a similar set of shared dictionary elements. The shared dictionary not only aids in aligning semantics across modalities but also functions as fine-grained "anchor points" to bridge the unaligned modalities. Consequently, by explicitly building fine-grained cross-modality alignment, the DHMD framework not only optimizes each modality to the fullest extent but also activates subtle emotion cues in weaker modalities so as to improve the performance of MER.


The contributions of this work can be summarized as:
\begin{itemize}
	\item We propose a unified multimodal KD framework, termed Decoupled Hierarchical Multimodal Distillation (DHMD), designed to facilitate dynamic and flexible distillation across various modalities for robust MER. To the best of our knowledge, this is the first approach that constructs multimodal alignment and coordination from both coarse- and fine-grained perspectives, offering a novel strategy for enhancing multimodal KD.

	\item In DHMD, we explicitly decouple multimodal representations into modality-irrelevant and modality-exclusive spaces to enable KD across these two decoupled spaces. For coarse-grained KD, DHMD employs a graph-based distillation mechanism that facilitates flexible knowledge transfer, with distillation directions and weights being automatically learned. For fine-grained KD, a shared dictionary is introduced, serving as a common semantic space to align and coordinate heterogeneous modalities. This hierarchical multimodal KD framework effectively aligns different modalities and leverages their intrinsic complementary properties.
	
	\item We conduct comprehensive experiments on four public MER datasets and obtain superior or comparable results than the state-of-the-arts. The experiments further verify that the two KD mechanisms consistently collaborate and effectively contribute to enhanced MER performance. Visualization results verify the feasibility of DHMD: both the graph edges and the dictionary activations exhibit meaningful distributional patterns in each of the decoupled multimodal feature spaces.
\end{itemize}

As an extension of our preliminary work in~\cite{li2023decoupled}, this paper makes the following improvements: 
1) We propose a unified multimodal distillation framework, i.e., DHMD, which concurrently integrates both coarse- and fine-grained KD mechanisms to effectively address the multimodal semantic gap inherent across different modalities. On the representative MER benchmarks, DHMD consistently shows its superiority.
2) In our proposed DHMD, we incorporate a shared dictionary to force the representations from different modalities to have a similar distribution over the discrete embedding space, which further consolidates the multimodal alignment and improves cross-modal consistency.
3) We provide more comprehensive experimental validation and analysis, including ablation studies that assess the contributions of both the coarse-grained and fine-grained KD mechanisms within DHMD. Our experiments also cover unimodal accuracy with and without
the two-stage KD components, along with comprehensive MER evaluations on two additional datasets, MUStARD~\cite{castro-etal-2019-towards} and UR-FUNNY~\cite{hasan-etal-2019-ur}, to further validate DHMD’s effectiveness across a variety of contexts.
4) We conduct thorough visual analysis to improve MER interpretability and confirm the efficacy of DHMD, e.g., we present visual analyses that illustrate the dynamic
graph edges with and without the two-stage KD mechanisms across the training process, as well as the activations within the cross-modal dictionary and the cross-modal attention matrices. These visualizations reveal DHMD’s internal processes, verifying the functionality of its hierarchical knowledge transfer and confirming DHMD’s capacity to refine multimodal representation learning.

\section{Related work}
\subsection{Multimodal Emotion Recognition}
Multimodal emotion recognition (MER) aims to infer human sentiment from the language, visual and acoustic information embedded in the video clips.
The heterogeneity across modalities can provide various levels of information for MER.
The mainstream MER approaches can be divided into two categories: fusion strategy-based \cite{TFN,LMF,MFN} and cross-modal attention-based \cite{MulT,PMR,MICA}.

The former aims to design sophisticated multimodal fusion strategies to generate discriminative multimodal representations, e.g., Zadeh \textit{et al.} \cite{TFN} designed a Tensor Fusion Network (TFN) that can fuse multimodal information progressively.
However, the inherent heterogeneity and the intrinsic information redundancy across modalities hinder the fusion between the multimodal features.
Therefore, some work aims to explore the characteristics and commonalities of multimodal representations via feature decoupling to facilitate more effective multimodal representation fusion \cite{MFM,MISA,FDMER}.
Hazarika \textit{et al.} \cite{MISA} decomposed multimodal features into modality-invariant/-specific components to learn the refined multimodal representations. The decoupled multimodal representations reduce the information redundancy and provide a holistic view of the multimodal data.
Recently, cross-modal attention-based approaches have driven the development of MER  since they learn the cross-modal correlations to obtain the reinforced modality representation. A representative work is MulT~\cite{MulT}. This work proposes the multimodal transformer that consists of a cross-modal attention mechanism to learn the potential adaption and correlations from one modality to another, thereby achieving semantic alignment between modalities.
Lv \textit{et al.} \cite{PMR} designed a progressive modality reinforcement method based on \cite{MulT}, it aims to learn the potential adaption and correlations from multimodal representation to unimodal representation. Our proposed DHMD has an essential difference with the previous feature-decoupling methods~\cite{MFM,MISA,FDMER}  because DHMD addresses inherent cross-modal 
heterogeneity by employing dedicated GD-Units in each decoupled multimodal space. This design enables adaptive and effective cross-modal distillation, eliminating the need for manual adjustment of distillation weights or strengths.

\begin{figure*}[htb]
	\centering{\includegraphics[width=0.95\linewidth]{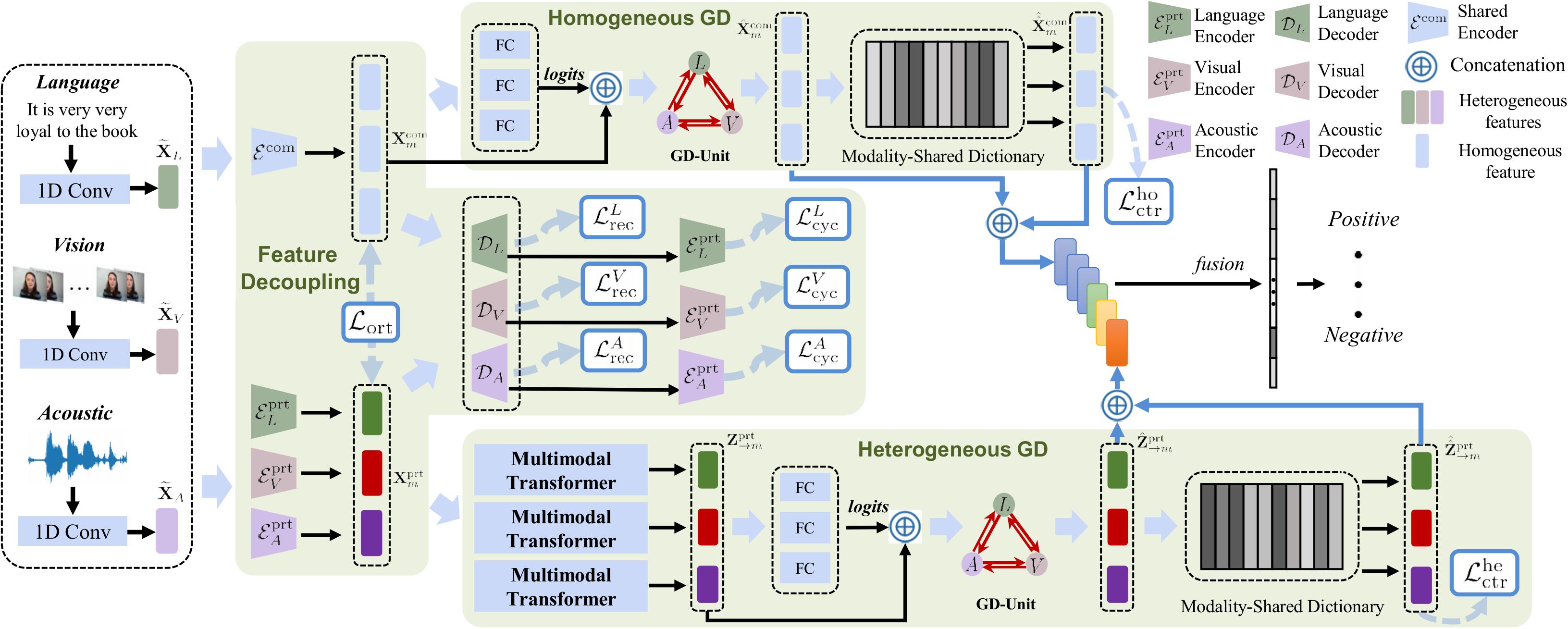}}
	\caption{The framework of DHMD. Given the input multimodal data, DHMD encodes their respective shallow features $\tbX_{m}$, where $m \in \{L, V, A\}$.
		In \textit{feature decoupling}, DHMD exploits the decoupled homo-/heterogeneous multimodal features 
		$\X^{\text{com}}_{m}$ / $\X^{\text{prt}}_{m}$ via the shared and exclusive encoders, respectively. $\X^{\text{prt}}_{m}$ will be reconstructed in a self-regression manner (Sec.~\ref{sec:decoupling}). 
		For \textbf{coarse-grained KD}, $\X^{\text{com}}_{m}$ and $\X^{\text{prt}}_{m}$ will be fed into a GD-Unit for adaptive KD in HoGD and HeGD, respectively (Sec.~\ref{sec:distillation}).
		For \textbf{fine-grained KD}, DHMD utilizes distinct dictionaries within each enhanced feature space to unify semantic granularities across modalities and achieve semantic alignment (Sec.~\ref{sec:cross-modal_CM}).
		Finally, the features from the two-stage KD mechanisms are adaptively fused for MER. 
	}
	\label{fig:framework}
\end{figure*}

\subsection{Knowledge Distillation and Multimodal Dictionary Learning}
\textbf{Knowledge Distillation}. The concept of KD was first proposed in \cite{KD} to transfer knowledge from teachers to students via minimizing the KL-Divergence between their prediction logits. Subsequently, various KD methods have been proposed \cite{logitsKD1,logitsKD2,logitsKD3,logitsKD4} based on \cite{KD} and further extended to distillation between intermediate features \cite{featKD1,featKD2,featKD3,featKD4}.

Most KD methods focus on transferring knowledge from the teacher to the student, while some recent studies have used graph structures to explore the effective message passing mechanism between multiple teachers and students with multiple instances of knowledge \cite{GDIJCAI,GDECCV,GDACCV}.
Zhang \textit{et al.} \cite{GDIJCAI} proposed a graph distillation (GD) method for video classification, where each vertex represented a self-supervised teacher and edges represented the direction of distillation from multiple self-supervised teachers to the student.
Luo \textit{et al.} \cite{GDECCV} considered the modality discrepancy to incorporate privileged information from the source domain and modeled a directed graph to explore the relationship between different modalities. Each vertex represented a modality and the edges indicated the connection strength (i.e., distillation strength) between one modality and another.
Different from them, we aim to use exclusive GD-Units in each of the decoupled feature spaces to facilitate effective cross-modality distillation.

\textbf{Multimodal Dictionary Learning}. Representing latent representation via discrete factorized elements has been explored for a long time. More methods can be roughly divided into Vector-Quantization-based~\cite{van2017neural} or dictionary-based~\cite{huang2021seeing, duan2022multi, chen2023revisiting}. For the former, the Vector Quantised Variational AutoEncoder (VQ-VAE)~\cite{van2017neural} first verified that image information can be effectively encoded by discrete codebooks. Subsequently, VideoBERT~\cite{sun2019videobert} proposed to cluster video features into a discrete set for effective cross-modal pretraining on raw video and language data. Li \textit{et al.}~\cite{li-etal-2022-unimo} proposed to conduct grounded learning on both images and texts via a sharing grounded space, which helps bridge unaligned images and texts, and align the visual and textual semantic spaces on different types of corpora. Unlike these methods, which primarily focus on decomposing input modalities into discrete tokens, our proposed DHMD leverages a shared dictionary to facilitate modality alignment and activate subtle emotional cues, particularly in the weaker modalities.

For dictionary-based multimodal feature fission,  Huang \textit{et al.}~\cite{huang2021seeing} proposed to extract comprehensive and compact image features through a visual dictionary (VD) that facilitates cross-modal understanding. However, the one-stream vision-specific dictionary in ~\cite{huang2021seeing} may well not capture the subtle cues in other modalities. Duan \textit{et al.}~\cite{duan2022multi} proposed to encode the two modalities into a joint vision-language coding space spanned by a dictionary of cluster centers. Chen \textit{et al.}~\cite{chen2023revisiting} proposed to embed both images and texts using shared finite discrete tokens and learn to encode the shared cross-modal semantic concepts. In comparison, the feature fission in our proposed DHMD is more intuitive, as we first build coarse-grained cross-modality alignment and then use the enhanced multimodal representations for feature fission to achieve fine-grained alignment. The hierarchical KD in our proposed method can be trained end-to-end and is modality-agnostic. More technical details 
can be found in Sec.~\ref{sec:cross-modal_CM}.

\section{Methodology}

The framework of our DHMD is illustrated in Fig.~\ref{fig:framework}.
It mainly consists of three parts: (1) \textbf{Multimodal feature decoupling}. Considering the significant distribution mismatch of modalities, we decouple multimodal representations into homogeneous and heterogeneous multimodal features through learning shared and exclusive multimodal encoders. The decoupling detail is introduced in Sec.~\ref{sec:decoupling}.
(2) \textbf{Coarse-grained KD via graph-distillation}. To facilitate a flexible knowledge transfer, we next distill the knowledge from homo-/hetero-geneous features, which are framed in two graph distillation units (GD-Unit), i.e., homogeneous GD (HoGD) and heterogeneous GD (HeGD).
In HoGD, homogeneous multimodal features are mutually distilled to compensate the representation ability for each other. In HeGD, multimodal transformers are introduced to explicitly build inter-modal correlations and semantic alignment for further distilling. The GD detail is introduced in Sec.~\ref{sec:distillation}. 
(3) \textbf{Fine-grained KD via cross-modal dictionary matching (DM)}. To further learn the coordinated multimodal space that captures fine-grained semantic relationships, we leverage a cross-modal DM mechanism that forces the decoupled
representations to have a similar distribution over the discrete embedding space (c.f. Sec.~\ref{sec:cross-modal_CM}).
Finally, the refined multimodal features through distilling are adaptively fused for robust MER. Below, we present the details of the three parts of DHMD.

\subsection{Multimodal feature decoupling}
\label{sec:decoupling}

We consider three modalities, i.e., \textit{language} (L), \textit{visual} (V), \textit{acoustic} (A).
Firstly, we apply individual 1D temporal convolutional layers for each modality, which serve to aggregate temporal information and extract low-level multimodal features, represented as: $\tbX_{m} \in \mathbb{R}^{T_{m} \times C_m}$, where $m \in \{L, V, A\}$ indicates a modality. Following this initial encoding process, each modality retains its original temporal dimension, allowing for the simultaneous management of both aligned and unaligned scenarios. Additionally, the feature dimensions across all modalities are standardized to be equal, i.e., $C_L= C_V = C_A = C$, facilitating efficient feature decoupling in subsequent stages.

To decouple the multimodal features into homogeneous component $\X_{m}^\sha$ and heterogeneous component $\X_{m}^\prt$, we employ a shared multimodal encoder, denoted as $\mathcal{E}^\sha$, alongside three modality-specific private encoders, $\mathcal{E}_{m}^\prt$. These encoders are utilized to explicitly predict the decoupled features. Formally,
\begin{equation}
	\X_{m}^\sha = \mathcal{E}^\sha(\tbX_{m}), \X_{m}^\prt=\mathcal{E}_{m}^\prt(\tbX_{m}).
\end{equation}
To differentiate between $\X_{m}^\sha$ and $\X_{m}^\prt$ and reduce feature ambiguity, we reconstruct the original coupled features $\tbX_m$ using a self-regression approach. Specifically, for each modality, we concatenate $\X_{m}^\sha$ and $\X_{m}^\prt$, and utilize a private decoder, $\mathcal{D}{m}$, to generate the coupled feature, represented as $\mathcal{D}{m}([\X_{m}^\sha, \X_{m}^\prt])$. This synthesized coupled feature is then re-encoded through the private encoders $\mathcal{E}_{m}^\prt$ to regress the heterogeneous features. Here, the notation $[.]$ denotes the concatenation of features. Formally, the discrepancy between the original and synthesized coupled multimodal features can be expressed as:
\begin{align}
	\mathcal{L}_{\text{rec}} = \| \tbX_m - \mathcal{D}_{m}([\X_{m}^\sha, \X_{m}^\prt]) \|_F^2.\label{equ:modality_rec}
\end{align}
Furthermore, the discrepancy between the vanilla and synthesized heterogeneous features can be mathematically expressed as:
\begin{align}
	\mathcal{L}_{\text{cyc}} = \| \X_{m}^\prt - \mathcal{E}_{m}^\prt(\mathcal{D}_{m}([\X_{m}^\sha, \X_{m}^\prt]))\|_F^2.
\end{align}

The reconstruction losses described above alone do not fully ensure feature decoupling. In practice, information may leak across representations, e.g., all modality-specific information could be encoded primarily in $\X_{m}^\prt$, allowing the decoders to easily synthesize the input, which would render the homogeneous multimodal features meaningless.

To strengthen feature decoupling, we posit that homogeneous representations derived from the same emotion but across different modalities should exhibit greater similarity than representations from the same modality but associated with different emotions. To enforce this criterion, we introduce a margin loss, defined as follows:

{\scriptsize
\begin{align}
	\mathcal{L}_{\text{mar}} =
	\frac{1}{|S|}\!\!\!\!\!\!\!\sum_{(i,j,k)\in S} \!\!\!\!\!\max(0, \alpha -\cos( {\X_{m[i]}^\sha, \X_{m[j]}^\sha} )\!\!+\!\!\cos( {\X_{m[i]}^\sha, \X_{m[k]}^\sha} ) ),
	\label{equ:homo_constraint}
\end{align}}

where we collect a triple tuple set $S=\{(i,j,k)| m[i]\neq m[j], m[i]= m[k], c[i]= c[j], c[i]\neq c[k] \}$. The $m[i]$ is the modality of sample $i$, $c[i]$ is the class label of sample $i$, and $\cos(\cdot,\cdot)$ means the cosine similarity between two feature vectors. The loss in Eq.~\ref{equ:homo_constraint} constrains the homogeneous features that belong to the same emotion but different modalities or vice versa to differ, and avoids deriving trivial homogeneous features. 
$\alpha$ is a distance margin. The distances of positive samples (same \textit{emotion}; different \textit{modalities})  are constrained to be smaller than that of negative samples (same \textit{modality}; different \textit{emotions}) by the margin $\alpha$.
To account for modality-irrelevant and modality-exclusive information, we introduce a soft orthogonality constraint to minimize redundancy between homogeneous and heterogeneous multimodal features:
\begin{equation}
	\mathcal{L}_{\text{ort}} = {\sum\limits_{m \in \{ L,V,A\}}{\cos(\X_{m}^\sha, \X_{m}^\prt)}}.
	\label{equ:orthogonality}
\end{equation}
Finally, we combine these constraints to form the decoupling loss,
\begin{align}
	\mathcal{L}_{\text{dec}} = \mathcal{L}_{\text{rec}} + \mathcal{L}_{\text{cyc}} + \gamma(\mathcal{L}_{\text{mar}} + \mathcal{L}_{\text{ort}}),
	\label{equ:feature_decoupling}
\end{align}
where $\gamma$ is the balance factor.


\subsection{Coarse-grained KD via Graph-Distillation Unit}
\label{sec:distillation}

For the decoupled homogeneous and heterogeneous multimodal features, we propose a graph distillation unit (GD-Unit) for each to perform adaptive KD. A typical GD-Unit consists of a directed graph $\mathcal{G}$, where each node $v_i$ corresponds to a modality, and $w_{i \rightarrow j}$ represents the distillation strength from modality $v_i$ to modality $v_j$. The distillation process from $v_i$ to $v_j$ is defined by the difference in their respective logits, denoted as $\epsilon_{i \rightarrow j}$. Let $\E$ represents the distillation matrix, where $E_{ij} = \epsilon_{i \rightarrow j}$. For a target modality $j$, the weighted distillation loss is formulated by considering the directed edges involved in the distillation process as follows:
\begin{equation}
	\zeta_{:j} = \sum_{v_i \in \mathcal{N}(v_j)} w_{i \rightarrow j} \times \epsilon_{i \rightarrow j},
\end{equation}
where $\mathcal{N}(v_j)$ represents the set of vertices injected to $v_j$.

To learn an adaptive weight that corresponds to the distillation strength $w$, we propose to encode the modality logits and the representations into the graph edges. Formally,
\begin{align}
	w_{i \rightarrow j}= g([[{f(\X_{i},\theta_1)}, {\X}_{i}], [{f(\X_{j},\theta_1)}, {\X_{j}}]], \theta_2),
	\label{equ:dynamic_edge}
\end{align}
where $[\cdot,\cdot]$ means feature concatenation, $g$ is a fully-connected (FC) layer with the learnable parameters $\theta_2$, and $f$ is a FC layer for regressing logits with the parameters $\theta_1$.   
The graph edge weights $\W$ with $W_{ij}=w_{i\rightarrow j}$ can be constructed and learned by repetitively applying Eq.~\ref{equ:dynamic_edge} over all pairs of modalities. To reduce the scale effects, we normalize $\W$ through the $softmax$ operation. Consequently, the graph distillation loss to all modalities can be written as:
\begin{align}
	\mathcal{L}_{\text{dtl}} = \|\W \odot \E\|_1,
	\label{equ:overall_distillation_loss}
\end{align}
where $\odot$ means element-wise product. Obviously, the distillation graph in a GD-Unit provides a base for learning dynamic inter-modality interactions. Meanwhile, it facilitates a flexible knowledge transfer manner where the distillation strengths can be automatically learned, thus enabling diverse knowledge transfer patterns.
Below, we elaborate on the details of HoGD and HeGD.

\textbf{HoGD.}
As illustrated in Fig.~\ref{fig:framework}, for the decoupled homogeneous features $\X_m^\sha$, as the distribution gap among the modalities is already reduced sufficiently, we input the features $\X_m^\sha$ and the corresponding logits $f(\X_m^\sha)$ to a GD-Unit and calculate the graph edge matrix $\W$ and the distillation loss matrix $\E$ according to Eq.~\ref{equ:dynamic_edge}. Then, the overall distillation loss $\mathcal{L}_{\text{dtl}}^{\text{homo}}$ can be naturally obtained via Eq.~\ref{equ:overall_distillation_loss}.

\textbf{HeGD.}
The decoupled heterogeneous features $\X_m^\prt$ focus on the diversity and the unique characteristics of each modality, and thus exhibit a significant distribution gap.
To mitigate this issue, we exploit the multimodal transformer~\cite{MulT} to bridge the feature distribution gap and build the modality adaptation.
The core of the multimodal transformer is the cross-modal attention unit ($\text{CA}$), which receives features from a pair of modalities and fuses cross-modal information. 
Take the language modality $\X_L^\prt$ as the source and the visual modality $\X_V^\prt$ as the target, the cross-modal attention can be defined as: $\Q_{V}=\X_V^\prt \P_{q}$, $\K_{L}=\X_L^\prt \P_{k}$, and $\V_{L}=\X_L^\prt \P_{v}$ where $\P_q, \P_k, \P_v$ are the learnable parameters. The individual attention head can then be expressed as:
\begin{align}
	\Z_{L \to V}^\prt&= \text{softmax}(\frac{\Q_{V}\K_{L}^{\top}}{\sqrt{d}})\V_L,
\end{align}
where $\Z_{L \to V}^\prt$ is the enhanced features from Language to Visual, $d$ means the dimension of $\Q_{V}$ and $\K_{L}$. For the three modalities in MER, each modality will be reinforced by the two others and the resulting features will be concatenated. For each target modality, we concatenate all enhanced features from other modalities to the target as the reinforced features, denoted with $\Z^\prt_{\rightarrow m}$, which are used in the distillation loss function as $\mathcal{L}_{\text{dtl}}^{\text{hetero}}$ that can be naturally obtained via Eq.~\ref{equ:overall_distillation_loss}.

\begin{figure}[htb]
	\centering{\includegraphics[width=1.0\linewidth]{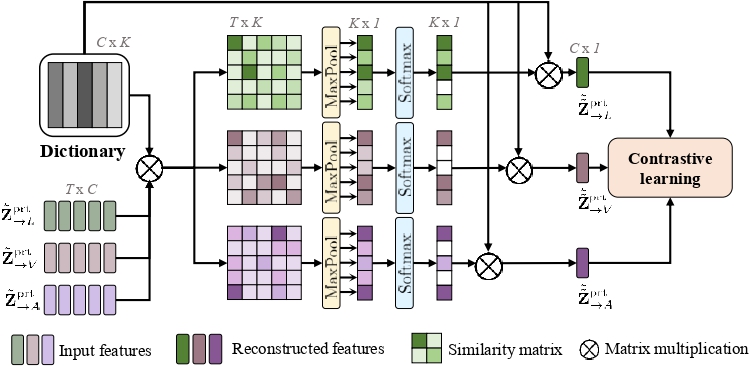}}
	\caption{Framework of the cross-modal Dictionary Matching (DM) mechanism for fine-grained KD. Decoupled multimodal features are projected onto a shared dictionary and subsequently reconstructed as weighted combinations of the dictionary elements. Overlapping elements within this dictionary capture the shared semantics across the modalities, facilitating fine-grained alignment.}
	\label{fig:fd_kd}
\end{figure}

\subsection{Fine-grained KD via Cross-Modal Dictionary Matching}
\label{sec:cross-modal_CM}
In addition to the coarse-grained KD, we further take as input the GD-enhanced multimodal representations for fine-grained semantic alignment, aiming to learn a shared and coordinated space that captures fine-grained semantic relationships between multimodal elements. The decomposed dictionary elements in one modality should share the same semantic meaning as elements in another modality. 

As to achieve this, We're introducing a cross-modal Dictionary Matching (DM) mechanism. This method forces the separate representations from different modalities to have a similar distribution in a discrete embedding space. As a result, the model can perceive emotion cues across modalities without direct supervision. We present the details of cross-modal DM in Fig.~\ref{fig:fd_kd}. Mathematically speaking, we define a shared dictionary (SD) as a matrix $\mathbf{D} \in \mathbb{R}^{K\times C}$, which consists of $K$ learnable embedding vectors with $C$-dim. The $k$-th embedding vector is denoted as $d_k$. With the shared dictionary $\mathbf{D}$, the decoupled multimodal representations are then projected to the SD space via computing the relevance between each unimodality and each embedding vector in $\mathbf{D}$. For clarity, the cross-modal DM  in the decoupled multimodal features are termed homogeneous DM (HoDM) and heterogeneous (HeDM). Here we take the HeDM as an example and present the details as follows.

Suppose $\tilde{\mathbf{{Z}}}^{\text{prt}}_{\to m} \in \mathbb{R}^{T\times C}$ denote the input heterogeneous features for cross-modal DM, where $T$ and $C$ denote the temporal and feature dimension, $m \in \{L, V, A\}$ indicates a modality. 
To get the element-wise association between $\tilde{\mathbf{{Z}}}^{\text{prt}}_{\to m}$ and the dictionary $\mathbf{D}$, we perform matrix multiplication operation $\mathbf{A}_m = \tilde{\mathbf{{Z}}}^{\text{prt}}_{\to m} \bigotimes \mathbf{D}^{tr}$, where $\mathbf{D}^{tr}$ means the transpose of $\mathbf{D}$.

The resultant  $\mathbf{A}_m^{ik} $ denotes the attention score between the $i$-th element  $[\tilde{\mathbf{{Z}}}^{\text{prt}}_{\to m}]_i$ and the $k$-th dictionary element $\mathbf{D}_k$. Subsequently, we perform $\mathtt{maxpool}$ operation column-wisely to acquire the overall attention score $\mathbf{a}_m$ between $\tilde{\mathbf{{Z}}}^{\text{prt}}_{\to m}$ and $\mathbf{D}_k$, as illustrated in Fig.~\ref{fig:fd_kd}. With the resultant attention score w.r.t each dictionary element, where each element may spot a certain word or facial frame, we are capable of better perceiving the crucial emotion cues in a more fine-grained manner.

Subsequently, we normalize the attention scores
via a $\mathtt{softmax}$ operation, which generates the final scores w.r.t each dictionary element:

\begin{align}
	\label{eq:attention_score}
	\alpha_m^{i}  =   \frac{exp(\mathbf{a}^i_m)}{\sum_{k=1}^{K} exp(\mathbf{a}^k_m)}.
\end{align}

Accordingly, the reconstructed modality representations can be expressed as:

\begin{align}
	\label{eq:soft_nearest_neighbor}
	\tilde{\tilde{\mathbf{{Z}}}}^{\text{prt}}_{\to m}  =   \sum_{1 \leq k \leq K} \alpha_{m}^k \mathbf{D}_k.
\end{align}

To effectively bring semantically corresponding modalities close together in a coordinated space and enforce strong equivalence between modality elements, the reconstructed modality representations are then incorporated for contrastive learning. We simply treat the representations from different modalities but same emotion category as positive pairs, and these from same modality but different emotion categories as negative pairs. We use $\mathcal{L}_{ctr}^{ho}$ and $\mathcal{L}_{ctr}^{he}$ to denote the learning target in the homogeneous and heterogeneous feature spaces, as illustrated in Fig.~\ref{fig:framework}. Formally, $\mathcal{L}_{ctr}^{he}$ can be formulated as:

{\footnotesize
\begin{align}
\mathcal{L}_{\text{ctr}}^{he} =
	\frac{1}{|U|}\!\!\!\sum_{(i,j,k)\in S}\!\!\!\!\!\max(0, \alpha\!\! -\!\! \cos( {\mathbf{z}_{m[i]}, \mathbf{z}_{m[j]}} )\!\!+\!\!\cos( {\mathbf{z}_{m[i]}, \mathbf{z}_{m[k]}} ) ), 
	\label{equ:hete_contrastive}
\end{align}}
where $\mathbf{z}_m = \tilde{\tilde{\mathbf{{Z}}}}^{\text{prt}}_{\to m} $ denotes the reconstructed modality features. For cross-modal DM objective in the homogeneous feature space, $\mathbf{z}_m = \tilde{\tilde{\mathbf{{X}}}}^{\text{com}}_{m} $.
We collect a triple tuple set within each online training batch. Note that $U=\{(i,j,k)| m[i]\neq m[j], m[i]= m[k], c[i]= c[j], c[i]\neq c[k] \}$. The $m[i]$ is the modality of sample $i$, $c[i]$ is the class label of sample $i$, and $\cos(\cdot,\cdot)$ means the cosine similarity between two feature vectors. With cross-modal DM, we are capable of automatically decomposing each modality into shared and discrete elements at an appropriate granularity, aiming to learn the fine-grained semantic correspondence across various modalities. For both the homogeneous and the heterogeneous feature spaces, cross-modal DM further enhances the discriminability w.r.t each modality. Besides, as for the heterogeneous feature space, cross-modal DM further bridges the semantic gap and will be clearly verified in Sec.~\ref{sec:ablation_study}.

\textbf{Feature fusion.} For heterogeneous features, we concatenate the coarse-grained and fine-grained KD-reinforced features, denoted as $\tilde{\mathbf{{Z}}}^{\text{prt}}_{\to m}$ and $\tilde{\tilde{\mathbf{{Z}}}}^{\text{prt}}_{\to m}$. A similar strategy is applied to the homogeneous features. These concatenated feature representations are subsequently utilized for MER. This approach allows the model to integrate information across different levels of granularity, enhancing its capacity for effective MER.

\subsection{Objective optimization}

We integrate the above losses to reach the full objective:
\begin{align}
	\mathcal{L}_{\text{total}} = \mathcal{L}_{\text{task}} + \lambda_1 \mathcal{L}_{\text{dec}} + \lambda_2\mathcal{L}_{{\text{dtl}}} + \lambda_3\mathcal{L}_{{\text{dic}}},
\end{align}
where $\mathcal{L}_{\text{task}}$ is the emotion task-related loss (here mean absolute error), $\mathcal{L}_{\text{dec}}$ means the feature decoupling loss in Eq.~\ref{equ:feature_decoupling}.
$\mathcal{L}_{\text{dtl}} = \mathcal{L}_{\text{dtl}}^{\text{homo}} + \mathcal{L}_{\text{dtl}}^{\text{hetero}}$ means the distillation losses generated by HoGD and HeGD.
$\mathcal{L}_{{\text{dic}}} = \mathcal{L}_{ctr}^{ho} + \mathcal{L}_{ctr}^{he}$ means the optimization target in the fine-grained KD part.
$\lambda_1, \lambda_2, \lambda_3$ control the importance of different constraints.

\section{Experimental results}
\textbf{Datasets.}
We evaluate DHMD on four representative MER datasets: CMU-MOSI \cite{mosi} and CMU-MOSEI \cite{mosei}, UR-FUNNY~\cite{hasan-etal-2019-ur}, and MUStARD~\cite{castro-etal-2019-towards} datasets. The experiments are conducted under the word-aligned  and unaligned settings for a more comprehensive comparison.
\textbf{CMU-MOSI} consists of 2,199 short monologue video clips. 
The acoustic and visual features in CMU-MOSI are extracted at a sampling rate of 12.5 and 15 Hz, respectively.
Among the samples, 1,284, 229 and 686 samples are used as training, validation and testing sets.

\textbf{CMU-MOSEI} contains 22,856 samples of movie review video clips from YouTube (approximately $10 \times$ the size of CMU-MOSI).
The acoustic and visual features were extracted at a sampling rate of 20 and 15 Hz, respectively.
According to the predetermined protocol,  16,326 samples are used for training, the remaining 1,871 and 4,659 samples are used for validation and testing. Each sample in CMU-MOSI and CMU-MOSEI was labeled with a sentiment score which ranges from -3 to 3, including \textit{highly negative}, \textit{negative}, \textit{weakly negative}, \textit{neutral}, \textit{weakly positive}, \textit{positive}, and \textit{highly positive}.
Following previous work \cite{PMR, MICA, MulT, MISA}, we evaluate the MER performance using the following metrics: 7-class accuracy (ACC$_{7}$), binary accuracy (ACC$_{2}$), F1 score, Mean Absolute Error (MAE) and Pearson Correlation (Corr).

\textbf{UR-FUNNY} serves as a widely recognized benchmark for evaluating humor detection in multimodal contexts. The dataset comprises punchlines extracted from TED talks, each annotated with a binary label to indicate whether humor is present as expressed by the speaker in the respective video. The dataset encompasses 1,866 videos featuring 1,741 distinct speakers. For this study, we employ an updated version of the dataset, which has undergone preprocessing to remove noisy and overlapping instances and includes additional contextual sentences. This refined dataset contains 16,514 utterances, divided into 10,598 for training, 2,626 for validation, and 3,290 for testing. Each sample in UR-FUNNY was annotated with a binary value where 1 demotes humor and 0 means non-humor. We use binary accuracy to represent the overall performance as that in ~\cite{hasan-etal-2019-ur}.
\textbf{MUStARD} is a multimodal video corpus designed for automated sarcasm detection. It comprises audiovisual utterances annotated with sarcasm labels and includes 690 video clips sourced from television shows. We utilize 539 utterances for training, 68 for validation, and 68 for testing. MUStARD is fundamentally a dialogue-level dataset, where the primary task is to classify the final utterance in a dialogue to determine whether the input modalities exhibit sarcasm. Each sample in the MUStARD dataset is annotated with a binary label, indicating whether the utterance is sarcastic (1) or non-sarcastic (0). We employ precision, recall, and F1-score as the evaluation metrics.

\textbf{Implementation details.} On the CMU-MOSI and CMU-MOSEI datasets, we extract the unimodal language features via GloVe \cite{GloVe} and obtain 300-dimensional word features.
For a fair comparison with~\cite{MISA, FDMER} under the aligned setting, we additionally exploit a BERT-base-uncased pre-trained model~\cite{bert} to obtain a 768-dimensional hidden state as the word features.
For visual modality, each video frame was encoded via Facet~\cite{Facet} to represent the presence of the total 35 facial action units~\cite{li2019self, li2020learning}. 
The acoustic modality was processed by COVAREP~\cite{COVAREP} to obtain the 74-dimensional features.  

For the UR-FUNNY and MUStARD datasets, the unimodal language features were extracted from a pre-trained BERT-base-uncased model~\cite{bert} to obtain a 768-dimensional hidden state as the word features. For visual modality, OpenFace facial behavioral analysis tool is used to extract the 371-dimensional facial expression features for each frame. The acoustic modality was processed by COVAREP~\cite{COVAREP} to obtain the 81-dimensional features. For MUStARD, the visual modality is processed with a pre-trained ResNet-152 model, which generates 2048-dimensional visual representations of each utterance. For the acoustic modality, the Librosa library~\cite{mcfee2015librosa} is utilized to extract 283-dimensional audio representations of each utterance.
 
The optimal setting for $\lambda_1$, $\lambda_2$, $\lambda_3$, $\gamma$ was set as 0.1, 0.05, 0.1, 0.1 via the MER performance on the validation set. The margin $\alpha$ was set as 0.1 for all the experiments.
We implemented all the experiments using PyTorch on a RTX 3090 GPU with 24GB memory. We set the training batch size as 16 and trained DHMD for 30 epochs until convergence.
Tab. \ref{tab:hyperparameters} illustrates the network architecture and the hyperparameter settings in DHMD. The GD-Units in \textit{HoGD} and \textit{HeGD} share the same architecture where the hidden dimension was set as 32.
Code as well as the pre-trained models will be publicly available. 


\begin{table}[h]
	\centering
	\caption{Hyperparameter settings in DHMD.  }\label{tab:hyperparameters}
	\scalebox{0.95}{
		\begin{tabular}{l|c|c|c|c}
			\hline
			Hyperparameter  &  [1] & [2] & [3] & [4]  \\
			\cline{2-3}
			\hline
			\hline
			\multicolumn{3}{c}{Feature decoupling}  \\
			\hline
			Kernel size for $\mathcal{C}_L$, $\mathcal{C}_V$, $\mathcal{C}_A$ & 5,5,5 & 5,3,3 & 5,5,5 & 5,3,5 \\
			Kernel size in $\mathcal{E}^\sha$ & 1 & 1 & 1 & 1\\
			Kernel size for $\mathcal{E}_{L}^\prt$, $\mathcal{E}_{V}^\prt$, $\mathcal{E}_{A}^\prt$  & 1,1,1 & 1,1,1 & 1,1,1 & 1,1,1 \\
			Kernel size for $\mathcal{D}_{L}$, $\mathcal{D}_{V}$, $\mathcal{D}_{A}$  & 1,1,1 & 1,1,1 & 1,1,1 & 1,1,1 \\
			\hline
			\hline
			\multicolumn{5}{c}{Coarse-grained KD: HoGD \& HeGD}  \\
			\hline
			Hidden dimension for a CA unit & 50 & 30 & 50 & 32\\
			Number of attention heads & 10 & 6 & 10 & 8 \\
			Layers of transformer  & 4 &4 & 4 & 2\\
			\hline
			\hline
			\multicolumn{5}{c}{Fine-grained KD: HoDM \& HeDM}  \\
			\hline
			Number of dictionary elements & 512 & 512 & 512 & 512 \\
			\hline
			\hline
	\end{tabular}}
\end{table}

\begin{table}[t]
	\centering
	\setlength{\tabcolsep}{1.5pt}
	\caption{Comparison on CMU-MOSI dataset. \textbf{Bold} is the best. \underline{Underline} denotes the second best.}\label{tab:MOSI}
		\begin{tabular}{c|c|ccccc}
			\hline
			Methods & Setting  &  ACC$_{7}$(\%) & ACC$_{2}$(\%) & F1(\%) & MAE & Corr \\
			\cline{3-5}
			\hline
			\hline
			EF-LSTM& \multirow{20}{*}{Aligned} &  33.7&  75.3&  75.2 & 1.023 & 0.608\\
			LF-LSTM &&  35.3& 76.8&76.7 & 1.015 & 0.625\\
			TFN \cite{TFN} & & 32.1& 73.9& 73.4 & 0.970 & 0.633\\
			LMF \cite{LMF} & & 32.8& 76.4& 75.7 & 0.912 & 0.668\\
			MFM \cite{MFM} & & 36.2& 78.1& 78.1 & 0.951 & 0.662\\
			RAVEN \cite{RAVEN} && 33.2& 78.0 & 76.6 & 0.915 & 0.691\\
			MCTN \cite{MCTN} && 35.6& 79.3& 79.1 & 0.909 & 0.676\\
			MulT \cite{MulT} & & 40.0&  83.0& 82.8 & 0.871 & 0.698\\
			PMR \cite{PMR} & & 40.6 & 83.6& 83.4 & -- & --\\
			
			MISA \cite{MISA}$^{*}$ & & 42.3& 83.4& 83.6 & 0.783 & 0.761\\
			FDMER \cite{FDMER}$^{*}$ & & 44.1& 84.6& 84.7 & 0.724 & 0.788\\
			IMDer \cite{wang2024incomplete}$^{*}$ & & 45.3 & 85.7 & 85.6 & 0.736 & 0.783\\
			GCNet \cite{lian2023gcnet}$^{*}$ & & 44.9& 85.2& 85.1 & -- & --\\
			CRIL  \cite{huang2023cross}$^{*}$ & & 46.3 & \underline{87.0} & \underline{86.9} & 0.695 & \textbf{0.812}\\
			TCTR~\cite{yang2026tctr}$^{*}$  & & 44.5 & 86.1 & 86.3 & 0.751 & 0.785\\
			EMOE~\cite{fang2025emoe}$^{*}$  & & \underline{47.7} & 85.4  & 85.4 & 0.710 & --\\
			DLF~\cite{wang2025dlf}$^{*}$  & & 47.1 & 85.1 & 85.0 & 0.731 & 0.781\\
			DEVA~\cite{wu2025enriching}$^{*}$  & & 46.3 & 86.3 & 86.3 & 0.730 & 0.787\\
			DMD~\cite{li2023decoupled}$^{*}$  & & 45.6 & 86.0  & 86.0 & \underline{0.691} & \underline{0.798}\\
			DHMD (\textbf{Ours})$^{*}$ & &\textbf{49.9} &\textbf{87.8}  &\textbf{87.7} & \textbf{0.683} & \textbf{0.812}\\
			\hline
			\hline
			EF-LSTM & \multirow{11}{*}{Unaligned} & 31.0 & 73.6 & 74.5 & 1.078 & 0.542\\
			LF-LSTM & & 33.7 & 77.6& 77.8 & 0.988 & 0.624\\
			RAVEN \cite{RAVEN} & &31.7 & 72.7 & 73.1 & 1.076 & 0.544\\
			MCTN \cite{MCTN} & & 32.7  & 75.9& 76.4 & 0.991 & 0.613\\
			MulT \cite{MulT} & &  39.1 & 81.1& 81.0 & 0.889 & 0.686\\
			PMR \cite{PMR} & & 40.6  & 82.4& 82.1 & -- & --\\
			MICA \cite{MICA}& &  40.8& 82.6& 82.7 & -- & --\\
			IMDer \cite{wang2024incomplete}$^{*}$ & & 44.5 & 85.2 & 85.2 & 0.744 & 0.771\\
			EMOE~\cite{fang2025emoe}$^{*}$  & & \underline{47.8} & \underline{85.4}  & \underline{85.3} & \underline{0.697} & --\\
			DMD \cite{li2023decoupled}$^{*}$  && 41.9 & 83.5  & 83.5 & 0.707 & \underline{0.786}\\
			DHMD (\textbf{Ours})$^{*}$ & &\textbf{48.4} &\textbf{86.9}  &\textbf{86.9} & \textbf{0.691} & \textbf{0.800}\\
			\hline
	\end{tabular}
	\begin{tablenotes}
		\centering
		\footnotesize
		\item[1] * means the input language features are BERT-based.
	\end{tablenotes}
\end{table}

\begin{table}[t]
	\centering
	\setlength{\tabcolsep}{1.5pt}
	\caption{Comparison on CMU-MOSEI dataset. \textbf{Bold} is the best. \underline{Underline} denotes the second best.}\label{tab:MOSEI}
		\begin{tabular}{c|c|ccccc}
			\hline
			Methods & Setting  &  ACC$_{7}$(\%) & ACC$_{2}$(\%) & F1(\%) & MAE & Corr\\
			\cline{3-5}
			\hline
			\hline
			EF-LSTM& \multirow{20}{*}{Aligned} &  47.4 &  78.2& 77.9 & 0.642 & 0.616\\
			LF-LSTM && 48.8 & 80.6 & 80.6 & 0.619 & 0.659\\
			Graph-MFN \cite{mosei} & & 45.0& 76.9&  77.0 & 0.710 & 0.540\\
			RAVEN \cite{RAVEN} && 50.0&  79.1&  79.5 & 0.614 & 0.662\\
			MCTN \cite{MCTN} && 49.6& 79.8& 80.6 & 0.609 & 0.670\\
			MulT \cite{MulT} & & 51.8& 82.5 & 82.3 & 0.580 & 0.703\\
			PMR \cite{PMR} & & 52.5& 83.3& 82.6 & -- & --\\
			
			MISA \cite{MISA}$^{*}$ & &  52.2&  85.5& 85.3 & 0.555 & 0.756\\
			FDMER \cite{FDMER}$^{*}$ & &  54.1& 86.1& 85.8 & 0.536 & 0.773\\
			IMDer \cite{wang2024incomplete}$^{*}$ & &  53.4 & 85.1 & 85.1 & 0.542 & 0.753\\
			GCNet \cite{lian2023gcnet}$^{*}$ & &  51.5 & 85.2 & 85.1 & -- & --\\
			CRIL \cite{huang2023cross}$^{*}$ & &  54.3 & \underline{86.6} & 86.5 & \textbf{0.525} & \underline{0.775}\\
			AcFormer \cite{zong2023acformer}$^{*}$ & &  \underline{54.7} & 86.5 & 85.8 & 0.531 & \textbf{0.786}\\
			TCTR~\cite{yang2026tctr}$^{*}$  & & 53.9 & 85.8 & 85.8 & 0.532 & 0.766\\
			EMOE~\cite{fang2025emoe}$^{*}$  & & 54.1 & 85.3 & 85.3 & 0.536 & --\\
			DLF~\cite{wang2025dlf}$^{*}$  & & 53.9 & 85.4 & 85.3 & 0.536 & 0.764\\
			DEVA~\cite{wu2025enriching}$^{*}$  & & 52.3 & 86.1 & 86.2 & 0.541 & 0.769\\
			DMD \cite{li2023decoupled}$^{*}$ & & 54.5 & \underline{86.6}  & \underline{86.6} & 0.537 & 0.773\\
			DHMD (\textbf{Ours})$^{*}$ & & \textbf{55.3} & \textbf{87.1}  &\textbf{87.0} & \underline{0.527} & \textbf{0.786}\\
			\hline
			\hline
			EF-LSTM& \multirow{11}{*}{Unaligned} &  46.3 & 76.1 & 75.9 & 0.680 & 0.585\\
			LF-LSTM &&  48.8  & 77.5&  78.2 & 0.624 & 0.656\\
			RAVEN \cite{RAVEN} & &45.5 & 75.4&75.7 & 0.664 & 0.599\\
			MCTN \cite{MCTN} & & 48.2  & 79.3& 79.7 & 0.631 & 0.645\\
			MulT \cite{MulT} & & 50.7  & 81.6& 81.6 & 0.591 & 0.694\\
			PMR \cite{PMR} & & 51.8  & 83.1& 82.8 & -- & --\\
			MICA \cite{MICA}& & 52.4  & 83.7& 83.3 & -- & --\\
			IMDer \cite{wang2024incomplete}$^{*}$ & &  53.0 & 84.7 & 84.7 & 0.582 & 0.726\\
			EMOE~\cite{fang2025emoe}$^{*}$  & & 53.9 & \underline{85.5}  & \underline{85.5} & \textbf{0.530} & --\\
			DMD \cite{li2023decoupled}$^{*}$ & & \underline{54.0} & \underline{85.5} & \underline{85.5} & \underline{0.556} & \underline{0.748}\\
			DHMD (\textbf{Ours})$^{*}$ & & \textbf{55.3} & \textbf{87.1}  &\textbf{87.0} & \textbf{0.530} & \textbf{0.776}\\
			\hline
	\end{tabular}
	\begin{tablenotes}
		\centering
		\footnotesize
		\item[1] * means the input language features are BERT-based.
	\end{tablenotes}
\end{table}

\begin{table}[t]
	\centering
	\setlength{\tabcolsep}{2.5pt}
	\caption{Comparison on MUStARD dataset. \textbf{Bold} is the best. \underline{Underline} denotes the second best.}\label{tab:MUStARD}
	\begin{tabular}{c|ccc}
		\hline
		Methods & Precision(\%) & Recall (\%) & F1 (\%)   \\
		\hline
		\hline
		SVM~\cite{castro-etal-2019-towards} &  71.9 &  71.4 & 71.5 \\
		MTL~\cite{chauhan2020sentiment} &  73.4 & 72.8 &   72.6 \\
		MulT \cite{MulT} & 74.8 & 74.6 &  74.2 \\
		IMDer \cite{wang2024incomplete} & 77.4 &  75.4 &  75.4 \\
		GCNet \cite{lian2023gcnet} & 77.5 & 76.1 & 76.1 \\
		DMD~\cite{li2023decoupled} &  \underline{77.8} & \underline{76.8} & \underline{76.9} \\
		DHMD (\textbf{Ours}) & \textbf{78.5} & \textbf{78.3} & \textbf{78.3} \\
			\hline
	\end{tabular}
\end{table}

\begin{table}[t]
	\centering
	\setlength{\tabcolsep}{2.5pt}
	\caption{Comparison on UR-FUNNY dataset. \textbf{Bold} is the best. \underline{Underline} denotes the second best.}
	\label{tab:UR-FUNNY}
	\begin{tabular}{c|ccc}
		\hline
		Methods &  ACC2 (\%)   \\
		\hline
		\hline
		C-MFN~\cite{hasan-etal-2019-ur} &  64.5 \\
		TFN~\cite{TFN}  &  64.7 \\
		LMF \cite{LMF} & 65.2 \\
		MFM \cite{MFM} & 65.2 \\
		SPT \cite{cheng2021multimodal} & 70.0 \\
		MulT~\cite{MulT} &  68.4 \\
		MISA~\cite{MISA}  &  \underline{70.6} \\
		IMDer~\cite{wang2024incomplete} &  69.1 \\
		GCNet~\cite{lian2023gcnet} &  69.2 \\
		DMD~\cite{li2023decoupled}  &  69.5 \\
		DHMD (\textbf{Ours}) &  \textbf{71.3} \\
		\hline
	\end{tabular}
\end{table}

\begin{table}[t]
	\centering
	\caption{Ablation study of the key components in DHMD. \textbf{Bold} denotes the best. \underline{Underline} denotes the second best.}
	\label{tab:Ablation}
		\begin{tabular}{c|cccc|ccc}
			\hline
			Dataset & FD & CA & GD & DM  & ACC$_{7}$ & ACC$_{2}$ & F1 \\
			\hline
			\hline
			\multirow{6}{*}{CMU-MOSI} & $\checkmark$ & $\checkmark$ & $\checkmark$ &  $\checkmark$ & \textbf{49.9} & \textbf{87.8} & \textbf{87.7}\\
			& $\checkmark$ & $\checkmark$ & $\checkmark$ &  $\times$ & 45.6 & 86.0 & 86.0\\
			& $\checkmark$ & $\checkmark$ & $\times$ &  $\checkmark$ & \underline{46.0} & \underline{86.4} & \underline{86.3}\\
			& $\checkmark$ & $\checkmark$ & $\times$ &  $\times$ & 45.0 & 85.2 & 85.3\\
			& $\times$ & $\checkmark$ & $\times$ &  $\checkmark$ & 44.8 & 85.0 & 84.8\\
			& $\checkmark$ & $\times$ & $\times$ &  $\times$ & 42.4 & 83.1 & 83.1\\
			& $\times$ & $\times$ & $\times$ &  $\times$ & 37.9 & 80.5 & 80.5\\
			\hline
			\hline
			\multirow{6}{*}{CMU-MOSEI}  & $\checkmark$ & $\checkmark$ & $\checkmark$ &  $\checkmark$ & \textbf{55.3} & \textbf{87.1} & \textbf{87.0}\\
			& $\checkmark$ & $\checkmark$ & $\checkmark$ &  $\times$ & \underline{54.5} & \underline{86.6} & \underline{86.6}\\
			&  $\checkmark$ & $\checkmark$ & $\times$ &  $\checkmark$ & 54.1 & 85.9 & 85.8\\
			&  $\checkmark$ & $\checkmark$ & $\times$ &  $\times$ & 53.0 & 85.0 & 84.9\\
			&  $\times$ & $\checkmark$ & $\times$ &  $\checkmark$ & 53.0 & 85.1 & 85.1\\
			&  $\checkmark$ & $\times$ & $\times$ &  $\times$ & 51.4 &  82.7 & 82.9\\
			&  $\times$ & $\times$ & $\times$ &  $\times$ & 49.0 & 80.8 & 80.7\\
			\hline
	\end{tabular}
\end{table}

\begin{figure*}[htb]
	\centering{\includegraphics[width=0.9\linewidth]{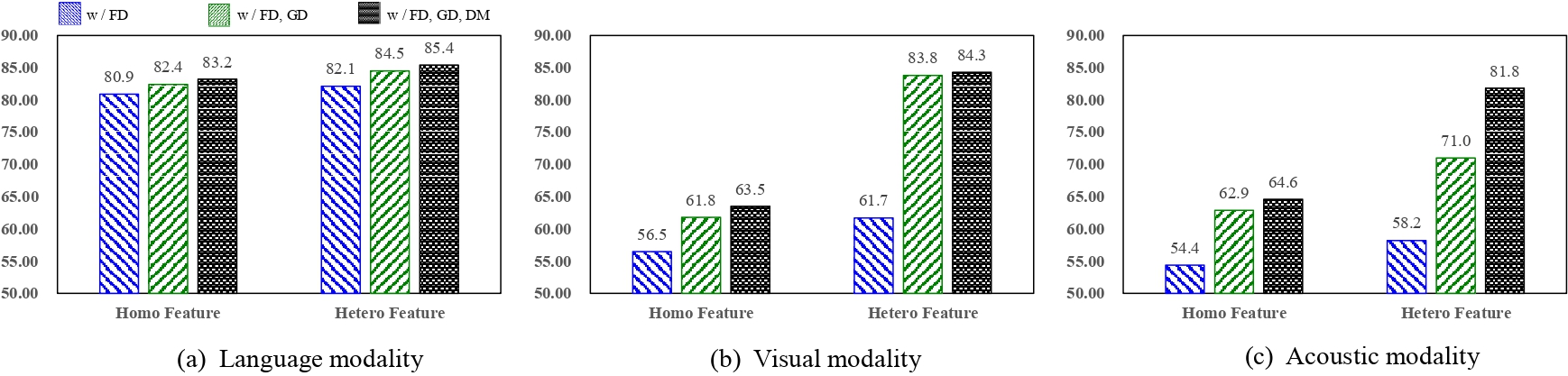}}
	\caption{Comparison of the decoupled homogeneous and heterogeneous features on \textit{CMU-MOSEI} dataset. \underline{w/ FD} means the baseline method where we merely obtain the decoupled features. \underline{w/ FD, GD} means adding a GD-Unit on each decoupled feature space based on \underline{w/FD} method. \underline{w/ FD, GD, DM} involves adding the cross-modal dictionary matching (DM) mechanism in each feature space based on \underline{w/ FD, GD} method. Evidently, both the coarse-grained KD mechanism, enhanced by graph distillation, and the fine-grained KD mechanism, based on dictionary matching, contribute consistently to cross-modal alignment.
	}
	\label{fig:single_modality_mosei}
\end{figure*}

\begin{table}[htb]
	\begin{center}
		\caption{Unimodal accuracy comparison on CMU-MOSEI dataset.}\label{tab:fd}
		\begin{tabular}{c|c|c}
			\hline
			\multirow{2}{*}{Methods} &
			w/o FD & w/ FD\\
			\cline{2-3}
			& Acc$_{2}$ (\%)~/~F1 (\%) & Acc$_{2}$ (\%)~/~F1 (\%)   \\
			\hline
			\hline
			$L$ only &81.2~~/~~81.4 & \textbf{82.7~~/~~82.5}
			\\
			$V$  only & 58.2~~/~~52.2& \textbf{62.8~~/~~60.0}
			\\
			$A$  only & 53.4~~/~~54.0 & \textbf{64.9~~/~~62.5}
			\\
			\hline
			Mean & 64.3~~/~~62.5& \textbf{70.1~~/~~68.3}
			\\
			STD & 12.1~~/~~13.4& \textbf{8.9~~/~~10.1}
			\\
			\hline
		\end{tabular}
	\end{center}
\end{table}

\subsection{Comparison with the state-of-the-art}
We compare DHMD with the current state-of-the-art MER methods under the same dataset settings (unaligned or aligned), including EF-LSTM, LF-LSTM, TFN~\cite{TFN}, LMF~\cite{LMF}, MFM~\cite{MFM}, RAVEN~\cite{RAVEN}, Graph-MFN~\cite{mosei}, MCTN~\cite{MCTN}, MulT~\cite{MulT}, PMR~\cite{PMR}, MICA~\cite{MICA}, MISA~\cite{MISA}, and FDMER~\cite{FDMER},  CRIL~\cite{huang2023cross}, AcFormer~\cite{zong2023acformer}, IMDer~\cite{wang2024incomplete}, GCNet~\cite{lian2023gcnet}, TCTR~\cite{yang2026tctr}, EMOE~\cite{fang2025emoe}, DLF~\cite{wang2025dlf} and
DEVA~\cite{wu2025enriching}.
IMDer and GCNet are originally designed for robust MER under missing modalities, which aim to explicitly recover missing data to facilitate MER. Both IMDer and GCNet have reported the numeric results in the complete modality setting for the CMU-MOSI and CMU-MOSEI datasets and provide publicly available codes. Therefore, we can seamlessly extend them to sarcasm and humor detection without modifying the algorithms due to the same modality input format (i.e., sequence feature). Specifically, under the complete modality setting, IMDer employs the multimodal transformers to fuse the multimodal data and predict the emotions. GCNet utilizes the graph neural network to fuse multimodal data and predict the target emotions.

\textbf{Comparison on CMU-MOSI and CMU-MOSEI datasets}. Tab.~\ref{tab:MOSI} and Tab.~\ref{tab:MOSEI} illustrate the comparison on CMU-MOSI and CMU-MOSEI datasets, respectively.
Obviously, our proposed DHMD obtains superior MER accuracy than other MER approaches under the unaligned and aligned settings.
Compared with the feature-disentangling-based MER methods~\cite{MISA, FDMER, MFM, wang2025dlf}, our proposed DHMD obtains consistent improvements, indicating the feasibility of the incorporated GD-Unit and cross-modal DM, which is capable of perceiving the various inter-modality dynamics. For a further investigation, we will visualize the learned graph edges and the dictionary activations in Sec.~\ref{sec:ablation_study}.
DHMD consistently outperforms the methods~\cite{MulT, MICA, PMR, zong2023acformer} that use multimodal transformer to learn cross-modal interactions and perform multimodal fusion. The reasons are two-fold: (1) DHMD takes the modality-irrelevant/-exclusive spaces into consideration concurrently and recudes the information redundancy via feature decoupling. (2) DHMD exploits hierarchical KD for coarse and fine grained semantic alignment. Both KD mechanisms are adaptive among various modalities. For our proposed DHMD, not only does the language modality enhance other modalities, other modalities can also enhance the language modality in an adaptive manner.

On the CMU-MOSEI dataset, Graph-MFN~\cite{mosei} illustrates unsatisfactory results because the heterogeneity and distribution gap across modalities hinder the learning of the modality fusion. As a comparison, the multimodal features in our proposed DHMD are decoupled into modality-irrelevant/-exclusive spaces.  For the latter space, we use multimodal transformer to bridge the distribution gap and align the high-level semantics, aiming to reduce the burden of absorbing knowledge from the heterogeneous features.
On both datasets, DHMD outperforms CRIL~\cite{huang2023cross} that aims to learn enhanced semantic representation learning of audio and video modalities with the guidance of textual tokens. The benefits of  DHMD are reasonable because all the modalities can benefit each other in the hierarchical KD training process. 
DHMD also outperforms DMD~\cite{li2023decoupled} that merely utilizes GD-Unit-empowered KD mechanism, indicating the superiority of incorporating cross-modal DM operation.
We will investigate and illustrate the contribution of each KD mechanism in Sec.~\ref{sec:ablation_study}.

\textbf{Comparison on UR-FUNNY and MUStARD datasets}. Sarcasm and humor have traditionally been most effectively conveyed through the language modality. However, in a typical multimodal context, accurately detecting sarcasm often relies on specific inter-modal cues to infer the speaker's intentions. For instance, sarcasm may manifest through a combination of verbal and non-verbal signals, such as tonal variations, exaggerated emphasis on certain words, elongated syllables, or even a neutral facial expression. As verified in Tab.~\ref{tab:UR-FUNNY} and Tab.~\ref{tab:MUStARD}, our proposed DHMD effectively addresses these requirements, consistently outperforming on both datasets. Through the hierarchical KD mechanism, which simultaneously enhances each modality and establishes inter-modal alignment, DHMD synchronizes across modalities over time, learning the intricate correlations among them.

In sarcasm detection, the input multimodal signals are often subtle, as the irony or incongruity is distributed across different modalities, making it challenging to effectively capture inter-modal cues. Multimodal sarcasm detection typically seeks to determine whether the literal expression is at odds with the speaker’s true intent within multimodal data. The incongruity of sarcasm often appears in highly fine-grained forms~\cite{lu2024fact}. Our proposed DHMD consistently outperforms prominent multimodal emotion recognition (MER) methods, such as Mult~\cite{MulT}, GCNet~\cite{lian2023gcnet}, IMDer~\cite{wang2024incomplete}, and DMD~\cite{li2023decoupled}. DHMD excels at capturing semantic-level sentiment incongruity through hierarchical knowledge transfer across each multimodal pair. We hypothesize that the combination of graph-empowered and dictionary-based KD effectively captures the intricate sarcasm-related associations and dependencies between different multimodal segments. Humor detection is known for its sensitivity to the distinctive characteristics of different modalities~\cite{hasan-etal-2019-ur}. These modality-specific dependencies are effectively captured by the representations within the DHMD framework, as demonstrated in Tab.~\ref{tab:UR-FUNNY}. Humor often intertwines with sentiment and is frequently expressed through exaggeration or irony, manifesting in modalities such as abrupt changes in tone or humorous gestures. The proposed DHMD framework is designed to capture these humor cues by leveraging feature disentanglement and hierarchical knowledge distillation mechanisms, enabling it to reliably identify both subtle and pronounced signals of humor.

\begin{figure*}[htb]
	\centering{\includegraphics[width=0.9\linewidth]{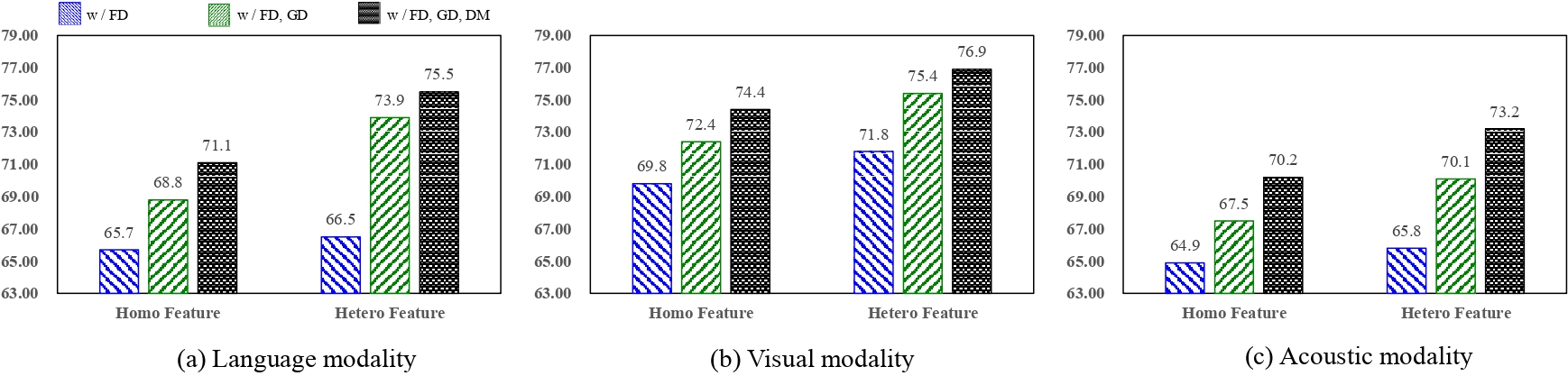}}
	\caption{Comparison of the decoupled homogeneous and heterogeneous features on \textit{MUStARD} dataset. Obviously, the coarse-/fine-grained KD mechanisms in our proposed DHMD collaboratively enhance the discriminability of each input modality. These mechanisms not only strengthen cross-modal alignment but also lead to improvements in the overall MER performance.
	}
	\label{fig:single_modality_MUStARD}
\end{figure*}

\begin{figure*}[htb]
	\centering{\includegraphics[width=0.9\linewidth]{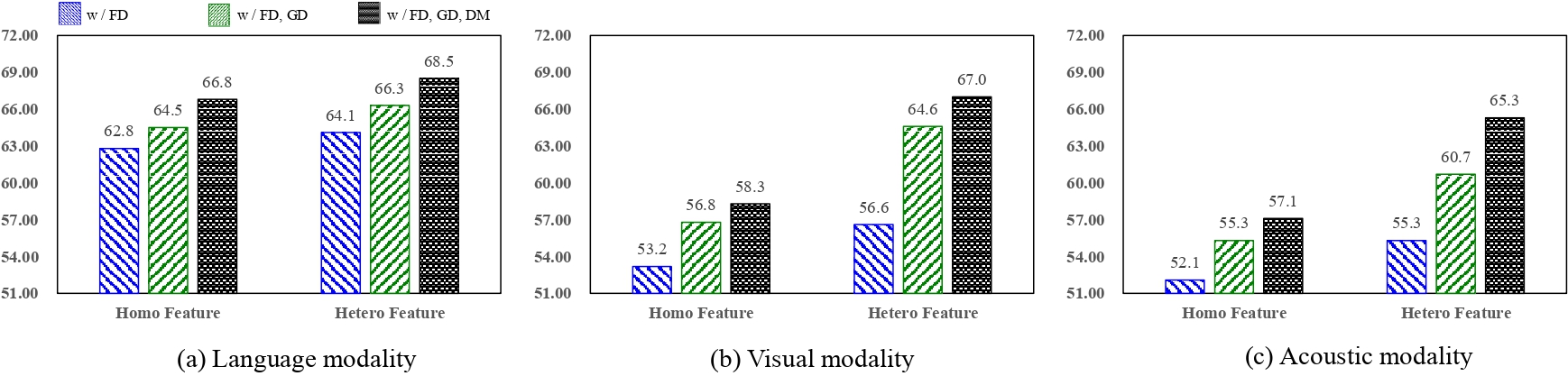}}
	\caption{Comparison of the decoupled homo-/hetero-geneous features on \textit{UR-FUNNY} dataset. Similar to Fig.~\ref{fig:single_modality_mosei} and Fig.~\ref{fig:single_modality_MUStARD}, our proposed DHMD automatically builds cross-modal alignment via the collaborative coarse and fine-grained KD mechanisms. 
	The discriminability w.r.t each unimodal has been effectively strengthed and the overall MER accuracy has also been consistently enhanced (c.f. Tab.~\ref{tab:UR-FUNNY}.) 
	}
	\label{fig:single_modality_UR-FUNNY}
\end{figure*}

\begin{table}[t]
	\centering
	\caption{Ablation study of graph distillation (GD) on MulT.}
	\label{tab:mult with gd}
		\scalebox{0.9} {\begin{tabular}{c|ccc|ccc}
			\hline
			\multirow{2}{*}{Methods} & \multicolumn{3}{c|}{CMU-MOSI} & \multicolumn{3}{c}{CMU-MOSEI}\\
			\cline{2-7} &
			ACC$_{7}$  & ACC$_{2}$ & F1 &
			ACC$_{7}$ & ACC$_{2}$ & F1   \\
			\hline
			\hline
			MulT & 39.1& 81.1 & 81.0 & 50.7 & 81.6 & 81.6 \\
			MulT (\textit{w/ GD}) & 39.4 & 82.2 & 82.2 & 51.0 & 82.3 & 82.5  \\
			MulT (\textit{w/ GD} \& \textit{DM}) & 40.7 & 82.6 & 82.5 & 52.7 & 83.7 & 83.8  \\
			DHMD (\textbf{Ours}) & \textbf{41.9} & \textbf{83.5} & \textbf{83.5} & \textbf{54.6} & \textbf{84.8} & \textbf{84.7}  \\
			\hline
	\end{tabular}}
\end{table}

\begin{figure*}[htb]
	\centering{\includegraphics[width=0.9\linewidth]{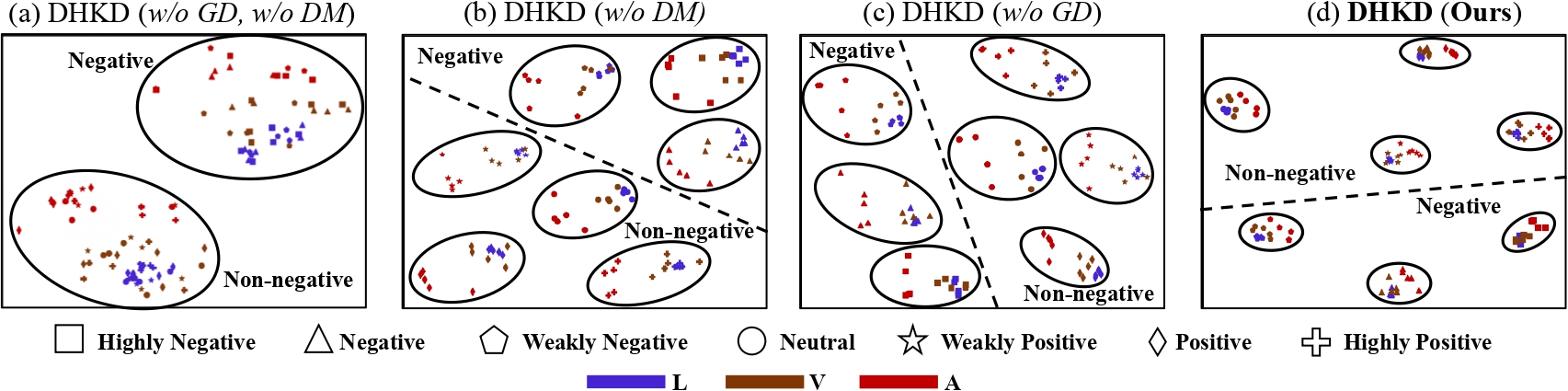}}
	\caption{t-SNE visualization of decoupled homogeneous space on \textit{CMU-MOSEI}  dataset. DHMD shows the promising emotion category (binary or 7-class)  separability in (d).}
	\label{fig:tsne-c}
\end{figure*}

\begin{figure*}[htb]
	\centering{\includegraphics[width=0.9\linewidth]{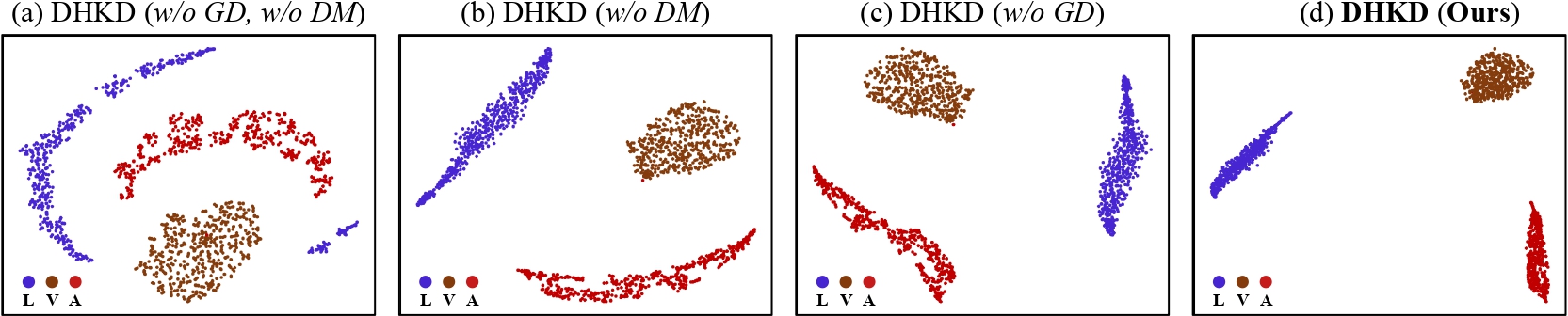}}
	\caption{Visualization of the decoupled heterogeneous features on \textit{CMU-MOSEI} dataset.
		DHMD shows the best modality separability in (d).
	}
	\label{fig:tsne-s}
\end{figure*}

\begin{figure*}[htb]
	\centering{\includegraphics[width=0.95\linewidth]{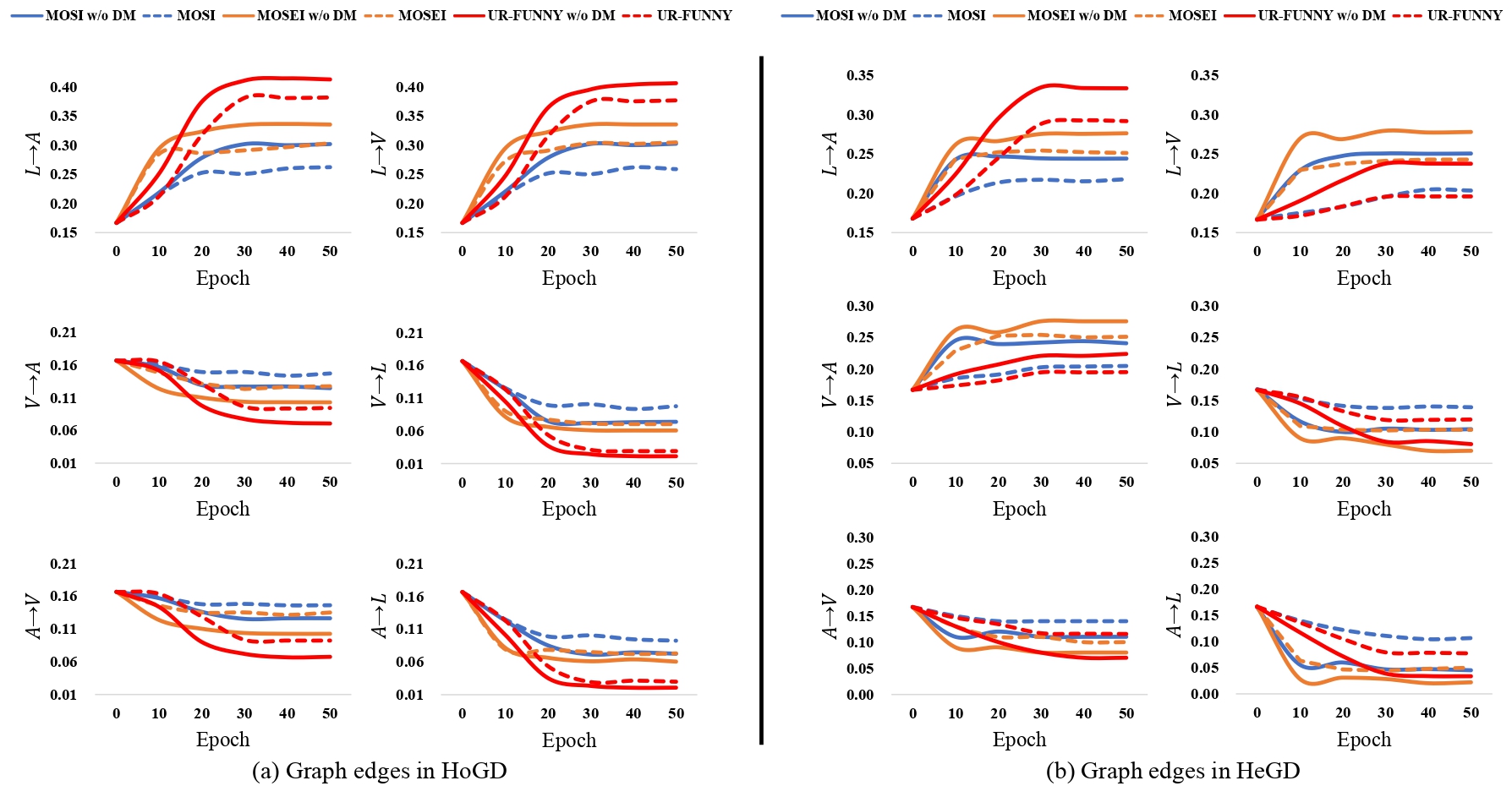}}
	\caption{Illustration of the graph edges in HoGD and HeGD w.r.t CMU-MOSI, CMU-MOSEI and UR-FUNNY datasets. 
	On the three datasets, language modality dominates.
		In (a), $L \to A$ and $L \to V$ are dominated because the homogeneous language features contribute most and the other modalities perform poorly.
		In (b), $L \to A$, $L \to V$, and $V \to A$ are dominated.  $V \to A$ emerges because the \textit{visual} modality enhanced its feature discriminability via the multimodal transformer mechanism in HeGD.
	}
	\label{fig:edge_mosei_mustard}
\end{figure*}

\begin{figure*}[htb]
	\centering{\includegraphics[width=0.95\linewidth]{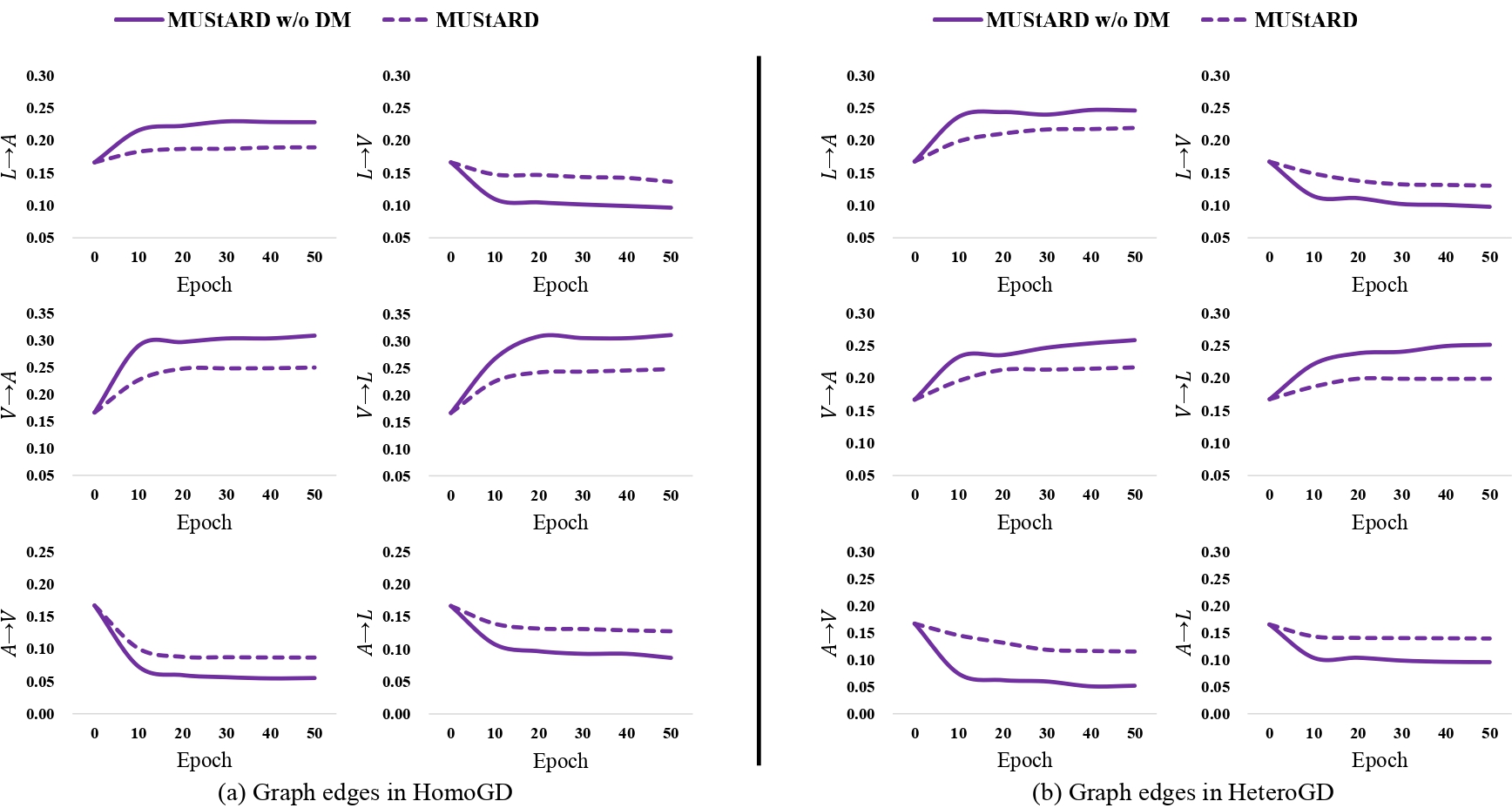}}
	\caption{Illustration of the graph edges on MUStARD dataset. Visual modality dominates.
		In both HoGD and HeGD, $V \to L$ and $V \to A$ and  $L \to A$ are dominated.
		In both the homogeneous and heterogeneous feature spaces, $L \to A$ emerges  because the language modality is notably enhanced (c.f. Fig.~\ref{fig:single_modality_MUStARD}). Besides, incorporating the cross-modal DM mechanism, our proposed DHMD consistently reinforces the distillation pathways with weaker weights, indicating cross-modal DM mitigates the multimodal distributional disparity during the alignment process.
	}
	\label{fig:edge_ur_funny}
\end{figure*}

\subsection{Analysis and discussion}
\label{sec:ablation_study}
\subsubsection{Quantitative analysis} 
We evaluate the effects of DHMD's key components, including feature decoupling (FD), graph distillation (GD), cross-modal attention unit (CA), cross-modal Dictionary Matching (DM). The results are illustrated in Tab.~\ref{tab:Ablation}. We conclude the observations below.

\textbf{Firstly}, FD enhances MER performance significantly, it indicates the decoupled and refined features can reduce information redundancy and provide discriminative multimodal features.
To further prove the effectiveness of FD, we conduct experiments on our baseline model with and without FD on CMU-MOSEI dataset. As shown in Tab.~\ref{tab:fd}, FD brings consistent improvements for each unimodality. Meantime, the performance gap for the three modalities is reduced as the standard deviations of ACC$_{2}$ and F1 are both decreased.

\textbf{Secondly}, combining FD with GD-Unit-based KD brings further benefits. This can be explained in two-fold: (1)
Although the homogeneous features are embedded in the same-dimension space, there still exists different discriminative capabilities for the modalities.
HoGD can improve the weak modalities through GD. To verify this, we conduct experiments with or without HoGD on CMU-MOSEI, MUStARD and UR-FUNNY datasets. The corresponding experimental results are shown in  Fig.~\ref{fig:single_modality_mosei}, Fig.~\ref{fig:single_modality_MUStARD}, and Fig.~\ref{fig:single_modality_UR-FUNNY}.  It is evident that the weaker modalities, such as the visual and acoustic modalities in the CMU-MOSEI and UR-FUNNY datasets, have shown obvious improvements with the application of HoGD. Similarly, although the weaker modalities in the MUStARD dataset are language and acoustic, comparable improvements are observed.
However, conducting HeGD without the cross-modal attention units will generate degraded performance, indicating the multimodal transformer plays a key role in bridging the multimodal distribution gap. Besides, with CA units and HeGD, the model will obtain conspicuous improvements, suggesting the importance of the taking advantage of the modality-exclusive features for robust MER.

\textbf{Thirdly}, unifying FD, GD, and DM enhances the MER accuracy consistently. This phenomenon suggests these three components function in a collaborative manner. Besides, we compare our proposed DHMD with the classical MulT~\cite{MulT} for further investigation. The results are shown in Tab.~\ref{tab:mult with gd}, where MulT (\textit{w/ GD}) means we add a GD-Unit on MulT to conduct adaptive knowledge transfer with the reinforced multimodal features. Essentially, the core difference between MulT (\textit{w/ GD} \& \textit{DM}) and DHMD is that DHMD incorporates feature decoupling.
The quantitative comparison in Tab.~\ref{tab:mult with gd} shows that DHMD obtains consistent improvements than MulT (\textit{w/ GD} \& \textit{DM}). It suggests decoupling the multimodal features before distillation is feasible and reasonable. Furthermore, DHMD achieves more pronounced improvements than the vanilla MulT, indicating the benefits of combining the feature decoupling and the hierarchical distillation mechanisms.

\subsubsection{Qualitative analysis} 
\textbf{Visualization of the decoupled features.}
We visualize the decoupled homogeneous and heterogeneous features of DHMD, DHMD (\textit{w/o GD}), DHMD (\textit{w/o DM}),  and DHMD (\textit{w/o GD, w/o DM}) in Fig.\ref{fig:tsne-c} and Fig.\ref{fig:tsne-s} for a quantitative comparison. 
DHMD (\textit{w/o GD, DM}) denotes DHMD without GD-Unit-based coarse-grained KD and also without DM-based fine-grained KD.
To visualize the homogeneous features, we randomly select 28 samples (four samples for each emotion category) in the test set of the CMU-MOSEI dataset.
For the heterogeneous features, we randomly select 400 samples in the test set of the CMU-MOSEI dataset.
The features of the selected samples are projected into a 2D space by t-SNE.

For the homogeneous multimodal features, the samples belonging to the same emotion category naturally cluster together due to their inter-modal homogeneity.
With the decoupled homogeneous features but without any  distillation mechanism in DHMD (\textit{w/o GD, w/o DM}), the samples merely show basic separability for the binary \textit{non-negative} and \textit{negative} categories. However, the samples are not distinguishable under the 7-class setting, indicating the features are not so discriminative than that of DHMD (\textit{w/o DM}) or DHMD (\textit{w/o GD}). The comparison between DHMD (\textit{w/o GD, w/o DM}), and DHMD (\textit{w/o DM}), and the comparison between DHMD (\textit{w/o GD, w/o DM}) and DHMD (\textit{w/o DM}) verifies the effectiveness of the graph distillation and dictionary-based KD on the homogeneous multimodal features.
Finally, our proposed DHMD illustrates the best feature separability w.r.t emotion categories, indicating the two KD mechanisms can promote each other effectively.

In the heterogeneous space, due to its inter-modal heterogeneity, the features of different samples should cluster by modalities.
As shown in Fig~\ref{fig:tsne-s}, DHMD shows the best feature separability, indicating the complementarity between modalities is mostly enhanced.
DHMD (\textit{w/o GD, w/o DM}), DHMD (\textit{w/o GD}) and DHMD (\textit{w/o DM}) show less feature separability than DHMD, indicating the importance of combing the coarse and fine grained hierarchical KD on the heterogeneous multimodal features.

\begin{figure}[htb]
	\centering{\includegraphics[width=0.95\linewidth]{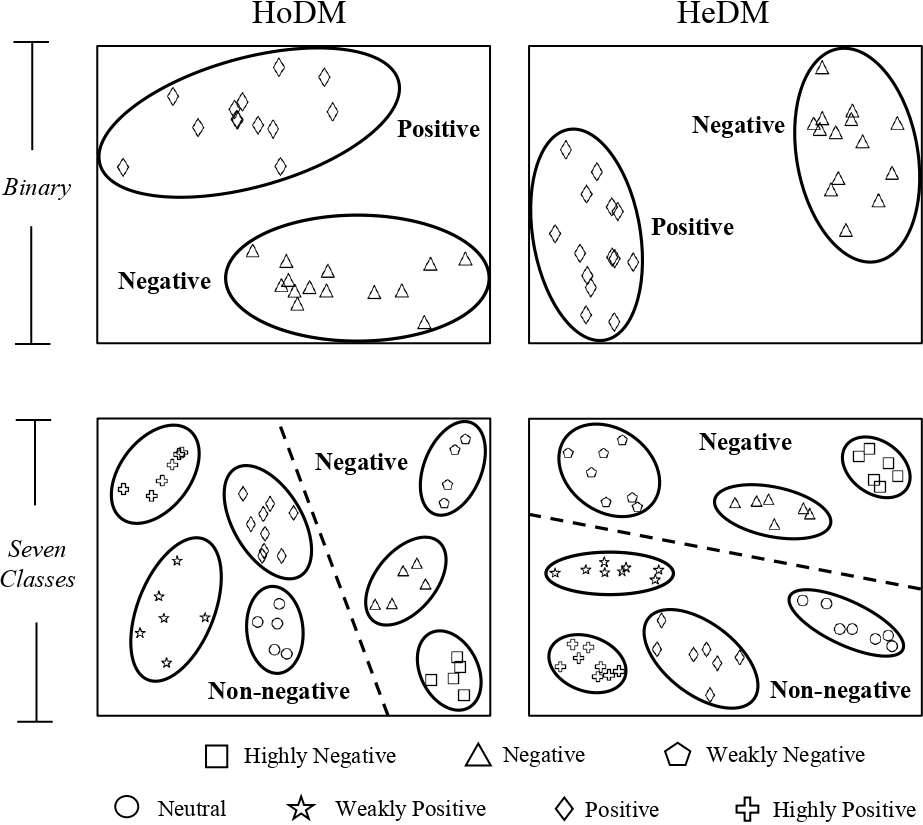}}
	\caption{t-SNE visualization of the cross-modal activated dictionary elements w.r.t each MER category.  The dictionary elements in the heterogeneous feature space display superior discriminative power compared to those in the homogeneous space. 
	}
	\label{fig:dictionary_elements}
\end{figure}

\begin{figure*}
	\begin{subfigure}
		\centering
		\includegraphics[width=\linewidth]{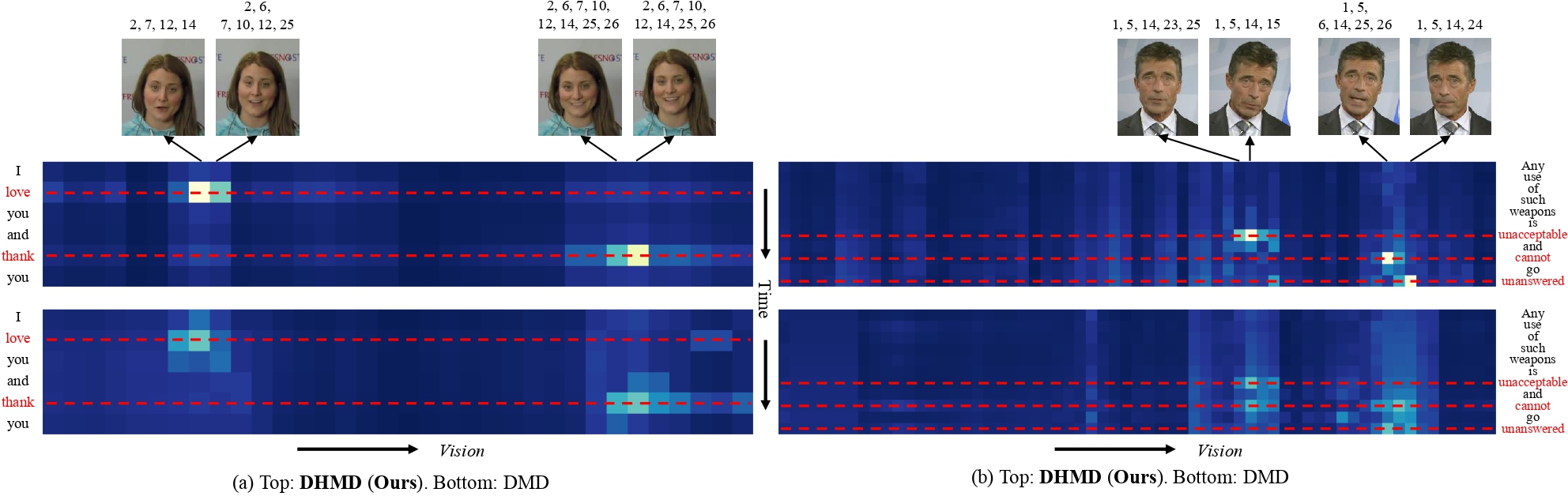}
		\label{fig:no_validation_1}
	\end{subfigure}%

	\begin{subfigure}
		\centering
		\includegraphics[width=\linewidth]{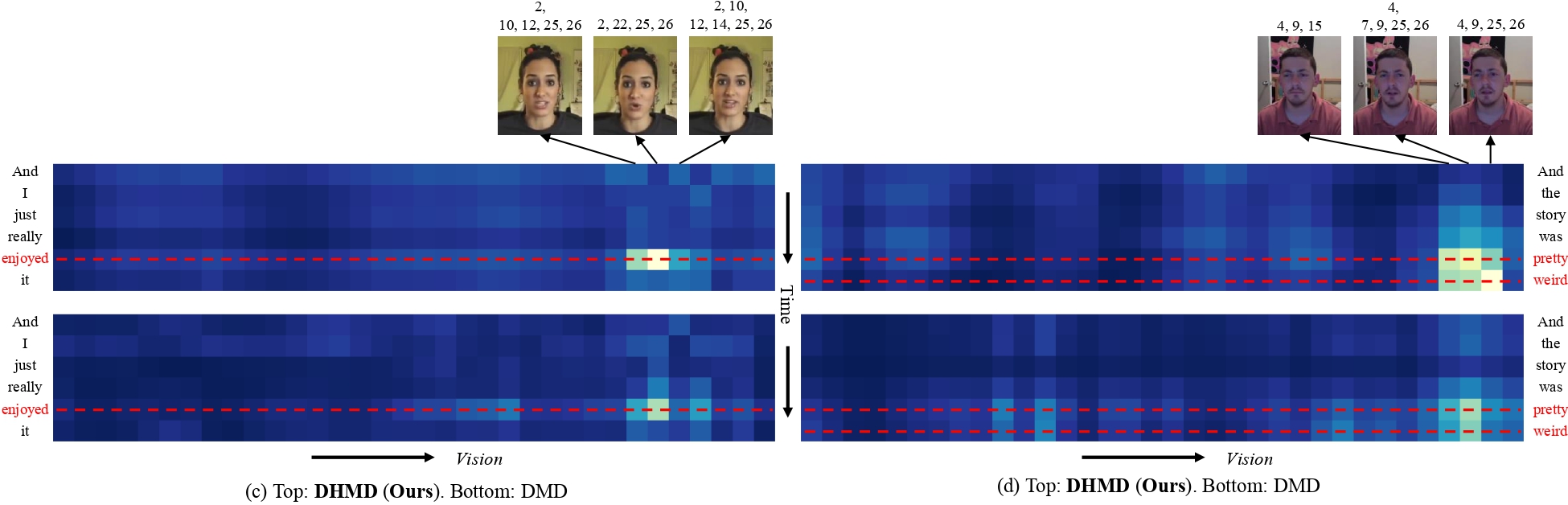}
		\label{fig:tcae_1}
	\end{subfigure}%
	\caption{Visualization of the cross-modal attention matrix activation for the proposed DHMD and DMD~\cite{li2023decoupled} on the CMU-MOSEI dataset. The language words that are closely related to emotion are highlighted in red. The numbers displayed above the video frames indicate the corresponding activated facial action units. In comparison to DMD, the proposed DHMD method captures more reliable correlations between elements across different modalities. Panels (a) and (c) represent a multimodal sample labeled as positive, while panels (b) and (d) depict one labeled as negative.}
	\label{fig:cross_transformer_attention}
\end{figure*}

\textbf{Visualization of graph edges in the GD-Units.} 
\textit{Driven by the quantitative improvements in Tab.\ref{tab:MOSI}, Tab.\ref{tab:MOSEI}, Tab.~\ref{tab:MUStARD}, and Tab.~\ref{tab:UR-FUNNY}, a question arises: \textit{How do the graph edges distribute during the training process? }}
We show the dynamic edges in each GD-Unit in Fig.~\ref{fig:edge_mosei_mustard} and Fig.~\ref{fig:edge_ur_funny} for analysis. Each graph edge corresponds to the strength of a directed distillation.

For Fig.~\ref{fig:edge_mosei_mustard}, we conclude the observations as: (1) The distillation in HoGD are mainly dominated by $L \to A$ and $L \to V$. This is because the decoupled homogeneous language modality still plays the most critical role and outperforms visual or acoustic modality with obvious advantages. 
As verified in Fig.~\ref{fig:single_modality_mosei}, for binary MER on the CMU-MOSEI dataset,  \textit{language}, \textit{visual}, \textit{acoustic} modality respectively obtains 80.9\%, 56.5\%, 54.4\% accuracy using the decoupled homogeneous features.  
(2) For HeGD,  $L \to A$, $L \to V$, and $V \to A$ are dominated. An interesting phenomenon is that $V \to A$ emerges.
This should be reasonable because the \textit{visual} modality enhanced its feature discriminability via the multimodal transformer mechanism in HeGD.
Actually, for CMU-MOSEI dataset, the binary accuracy w.r.t the three modalities (language, visual, acoustic) with or without HeGD are respectively: 82.1\% vs. 84.5\%, 61.7\% vs. 83.8\%, and 58.2\% vs. 71.0\%.
Similar phenomenons can be observed on UR-FUNNY and CMU-MOSI datasets (c.f. Fig.~\ref{fig:edge_mosei_mustard}).

For Fig.~\ref{fig:edge_ur_funny}, the distillation in HoGD are mainly dominated by $V \to L$, $V \to A$ and $L \to A$. 
For HeGD,  $L \to A$, $V \to A$, and $V \to L$ are dominated. These phenomenons can be further explained via Fig.~\ref{fig:single_modality_MUStARD} which shows the binary accuracy with respect to each modality on MUStARD dataset, both with and without the graph-empowered KD mechanism. Unlike the other three MER datasets where the language modality is predominant, the visual modality in MUStARD dataset  plays a more crucial role, surpassing the other two modalities. In both the homogeneous and heterogeneous feature spaces, the discriminative capacity of the language modality is notably enhanced, thus leading to the prominence of $L \to A$ interactions.

Additionally, when incorporating cross-modal DM mechanism, our proposed DHMD is capable of enhancing the cross-modal distillation strength whose vanilla distillation strength is low, e.g., $V \to A$, $V \to L$, $A \to V$, and $A \to L$ in HoGD, $V \to L$, $A \to V$, and $A \to L$ in HeGD (c.f. Fig.~\ref{fig:edge_mosei_mustard}). This phenomenon suggests that the cross-modal DM effectively enhances the efficiency of cross-modal distillation and promotes more comprehensive transfer of cross-modal knowledge. We posit that the cross-modal DM mitigates the distributional disparity between modalities during the alignment process, reinforces the distillation pathways with weaker weights, and consequently optimizes the overall utilization of graph distillation.

Conclusively, the learned graph edges in the GD-Unit represent the strength of the learned knowledge distillation from one modality to another. These edges are not static but are dynamically determined based on the specific input example, allowing for flexible and adaptive knowledge transfer. This dynamic structure overcomes the limitations of fixed-weight methods and helps the framework effectively address the inherent feature distribution mismatches and varying contributions of different modalities.

\textbf{Visualization of the dictionary activations in cross-modal DM.}
On the CMU-MOSEI dataset, we conduct a thorough investigation for the cross-modal activated dictionary elements w.r.t each category to analyze what exactly the dictionary has learned. We show the results in Fig.~\ref{fig:dictionary_elements}. The final selected dictionary elements demonstrate not only class discrimination but also cross-modal advantages. This is because the proposed DM mechanism is designed to serve the dictionary elements as a common semantic space, aligning different modalities within each of the decoupled feature spaces and unifying semantic granularities across various input modalities. As illustrated in Fig.~\ref{fig:dictionary_elements}, we have three observations: (1)  The dictionary elements in the heterogeneous feature space display superior discriminative power compared to those in the homogeneous space, consistent with the phenomenons in Fig.\ref{fig:single_modality_mosei}, Fig.~\ref{fig:single_modality_MUStARD}, and Fig.~\ref{fig:single_modality_UR-FUNNY}. This suggests that dominant discriminative features reside in modality-specific factors for each unimodal representation, further validating the feasibility of feature decoupling. (2) Within each MER category, the dictionary elements exhibit clear intra-category variation, indicating that the dictionary elements have become more representative of class diversity. This observation also suggests that the dictionary elements within a given class can effectively associate diverse inputs from different modalities with the same underlying concept and facilitate enhanced multimodal alignment.

Conclusively, the proposed DHMD framework incorporates a cross-modal DM mechanism that unifies the semantic granularities of different modalities within each decoupled MER space. This fine-grained alignment operation effectively bridges the inherent semantic gap between modalities and activates subtle emotion cues, particularly within weaker modalities, thus enhancing the overall MER performance.
 
\textbf{Visualization of attention maps in DHMD.} 
Fig.~\ref{fig:cross_transformer_attention} shows the visualizations of the cross-modal attention matrix in our proposed DHMD and DMD \cite{li2023decoupled} on the CMU-MOSEI dataset.
Obviously, our proposed DHMD is capable of perceiving more reliable correlations between the modalities.
As shown in Fig. \ref{fig:cross_transformer_attention} (a) and Fig. \ref{fig:cross_transformer_attention} (c), DHMD successfully correlates the positive words, e.g., ``love'' and ``enjoyed'', with the faces that show the corresponding action units, e.g., AU6 (Cheek Raiser), AU10 (Upper Lip Raiser), AU12 (Lip Corner Puller), AU25 (Lips part).
In Fig. \ref{fig:cross_transformer_attention} (b) and Fig. \ref{fig:cross_transformer_attention} (d), DHMD demonstrates the ability to establish reliable cross-modal correlations, even when faces in the visual modality exhibit few activated AUs. 
The visualizations in Fig.~\ref{fig:cross_transformer_attention} are reasonable because the features in DHMD are further refined and aligned via cross-modal DM mechanism. These results also indicate that fine-grained modality alignment is effective in capturing more subtle emotional cues across different modalities.
It is worth noting that the comparisons in Fig.~\ref{fig:cross_transformer_attention} are consistent with Tab.~\ref{tab:MOSI}$\sim$Tab.~\ref{tab:UR-FUNNY} where DHMD outperforms DMD on various MER datasets. 

\subsubsection{Parameter sensitivity analysis}
\begin{figure*}[htb]
	\centering{\includegraphics[width=0.9\linewidth]{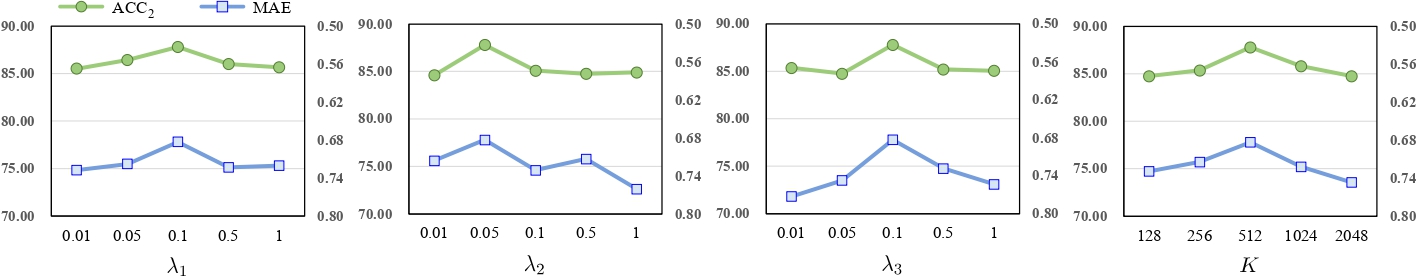}}
	\caption{ Parameter sensitivity analysis of $\lambda_1$, $\lambda_2$, $\lambda_3$ and dictionary size $K$ on CMU-MOSI dataset.}
	\label{fig:parameter_sensitivity}
\end{figure*}

Thank you for your suggestion. To provide comprehensive insight into the stability of our model, we have conducted a detailed parameter sensitivity analysis on the CMU-\linebreak MOSI dataset, focusing on two critical groups of hyper-\linebreak parameters: (1) Loss Balancing Weights: The weights $\lambda_1$, \linebreak $\lambda_2$ and $\lambda_3$, which govern the respective contributions of the feature decoupling, graph distillation, and dictionary matching mechanisms. (2) Dictionary Size: the parameter $K$, which controls the semantic granularity within our fine-grained knowledge distillation mechanism. The experimental results are illustrated in Fig.~\ref{fig:parameter_sensitivity}, from which we draw the following conclusions: (1) Robustness to Loss Weights: The performance of DHMD remains largely stable across a reasonable range of values around the default settings (e.g., $\lambda_1 \in 0.01, 0.05, 0.1, 0.5, 1.0$). This stability effectively demonstrates the robustness of our method to the balance of the multiple loss terms. (2) Optimal Dictionary Size: Performance improves progressively as $K$ increases from 128 to 512, confirming the necessity of a sufficient number of dictionary atoms to capture fine-grained semantic cues. However, further increasing $K$ to 1024 or 2048 results in slight performance degradation, likely due to increased sparsity. This confirms that $K = 512$ represents an optimal trade-off between representational capacity and computational efficiency.

\section{Conclusion and discussion}
Within this paper we have proposed a decoupled hierarchical multimodal distillation method (DHMD) for MER. 
Our method is inspired by the observation that the contribution of different modalities varies significantly. Therefore, robust MER can be achieved by distilling the reliable and generalizable knowledge across the modalities. To mitigate the modality heterogeneities, DHMD decouples the modal features into modality-irrelevant/-exclusive spaces in a self-regression manner. For coarse-grained KD, two GD-Units are incorporated for each decoupled feature to facilitate adaptive cross-modal  distillation.
Additionally, we propose a cross-modal dictionary matching mechanism for fine-grained KD and multimodal alignment. This mechanism automatically aligns the semantic granularities of various input modalities, resulting in more discriminative multimodal representations. Both quantitative and qualitative experiments consistently validate the effectiveness of DHMD. However, a limitation of our approach is its lack of integration with prevalent multimodal foundation models, which could potentially enhance knowledge transfer and result in more robust MER. We will explore this in future research.

\ifCLASSOPTIONcompsoc
\section*{Acknowledgments} This work was partly supported by the Natural Science Foundation of China (62576099, 62320106007, 62521007, 62276180), the Start-up Research Fund of Southeast University (RF1028625087), the RIE2020 AME Programmatic Fund, Singapore (No. A20G8b0102) and A*STAR Prenatal / Early Childhood Grant (No. H22P0M0002), the Big Data Computing Center of Southeast University.
\else
\section*{Acknowledgment}
\fi

\ifCLASSOPTIONcaptionsoff
  \newpage
\fi

\normalem
\bibliographystyle{IEEEtran} 
\bibliography{egbib}


\begin{IEEEbiography}[{\includegraphics[width=1in,height=1.25in,clip,keepaspectratio]{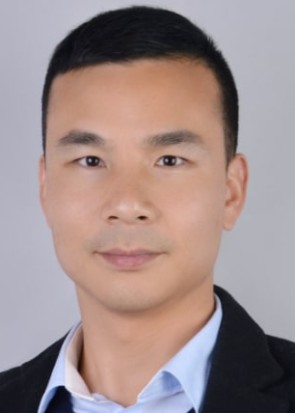}}]
	{Yong Li} received his Ph.D. in Computer Science from the Institute of Computing Technology (ICT) at the Chinese Academy of Sciences (CAS), Beijing, in July 2020. He served as an Assistant Professor at Nanjing University of Science and Technology from 2020 to 2022. Since 2023, he has been a Research Fellow in the Department of Computer Science and Engineering at Nanyang Technological University (NTU). His research focuses on deep learning, self-supervised learning, and affective computing.
\end{IEEEbiography}

\begin{IEEEbiography}[{\includegraphics[width=1in,height=1.25in,clip,keepaspectratio]{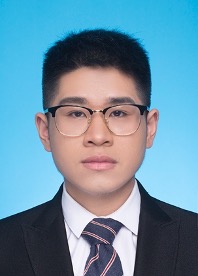}}]
	{Yuanzhi Wang} received the B.S. and master’s degrees from the Wuhan Institute of Technology, Wuhan, China, in 2019 and 2022, respectively. He is currently working toward the Ph.D. degree with the School of Computer Science and Engineering, Nanjing University of Science and Technology, Nanjing, China. His research interests include multimodal machine learning, generative model, and image/video processing.
\end{IEEEbiography}

\begin{IEEEbiography}
	[{\includegraphics[width=1in,height=1.25in,clip,keepaspectratio]{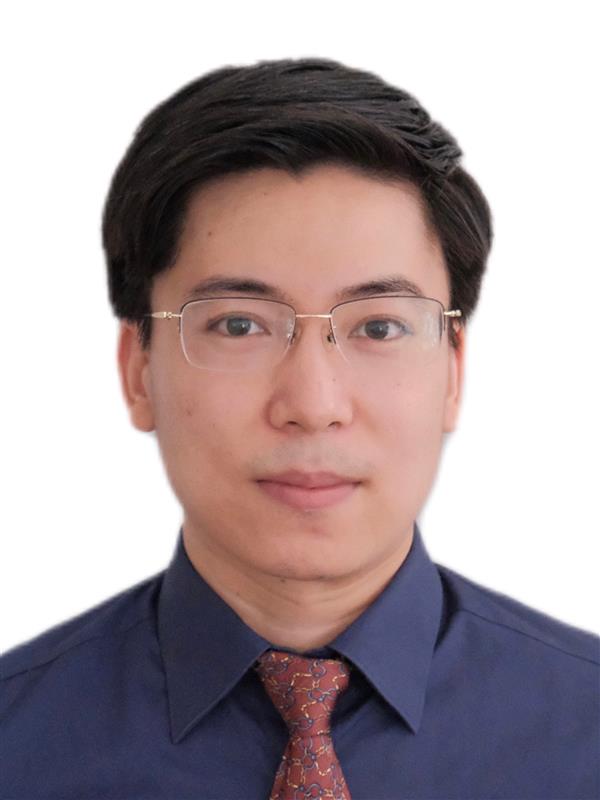}}]{Yi Ding}(Member, IEEE) earned a Ph.D. in Computer Science and Engineering from Nanyang Technological University, Singapore, in 2023, a master's degree in Electrical and Electronics Engineering from the same university in 2018, and a bachelor's degree in Information Science and Technology from Donghua University, Shanghai, China, in 2017. His research interests encompass brain-computer interface, deep learning, graph neural networks, neural signal processing, affective computing, and multimodal learning.
\end{IEEEbiography}

\begin{IEEEbiography}[{\includegraphics[width=1in,height=1.25in,clip,keepaspectratio]{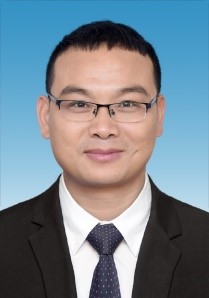}}]
	{Shiqing Zhang} (Member, IEEE) received the Ph.D. degree at school of Communication and Information Engineering, University of Electronic Science and Technology of China, in 2012. He was a postdoctor with the School of Electronic Engineering and Computer Science, Peking University, Beijing, China. Currently, he is a professor of the Institute of Intelligent Information Processing, Taizhou University, China. His research interests include machine learning, pattern recognition, computer vision, and affective computing. He has published over 70 papers in journals and conferences such as IEEE Transactions on Multimedia, IEEE Transactions on Circuits and Systems for Video Technology, IEEE Transactions on Affective Computing, and ACM MM. He serves as an Associate Editor for IEEE Transactions on Affective Computing.
\end{IEEEbiography}

\begin{IEEEbiography}[{\includegraphics[width=1in,height=1.25in,clip,keepaspectratio]{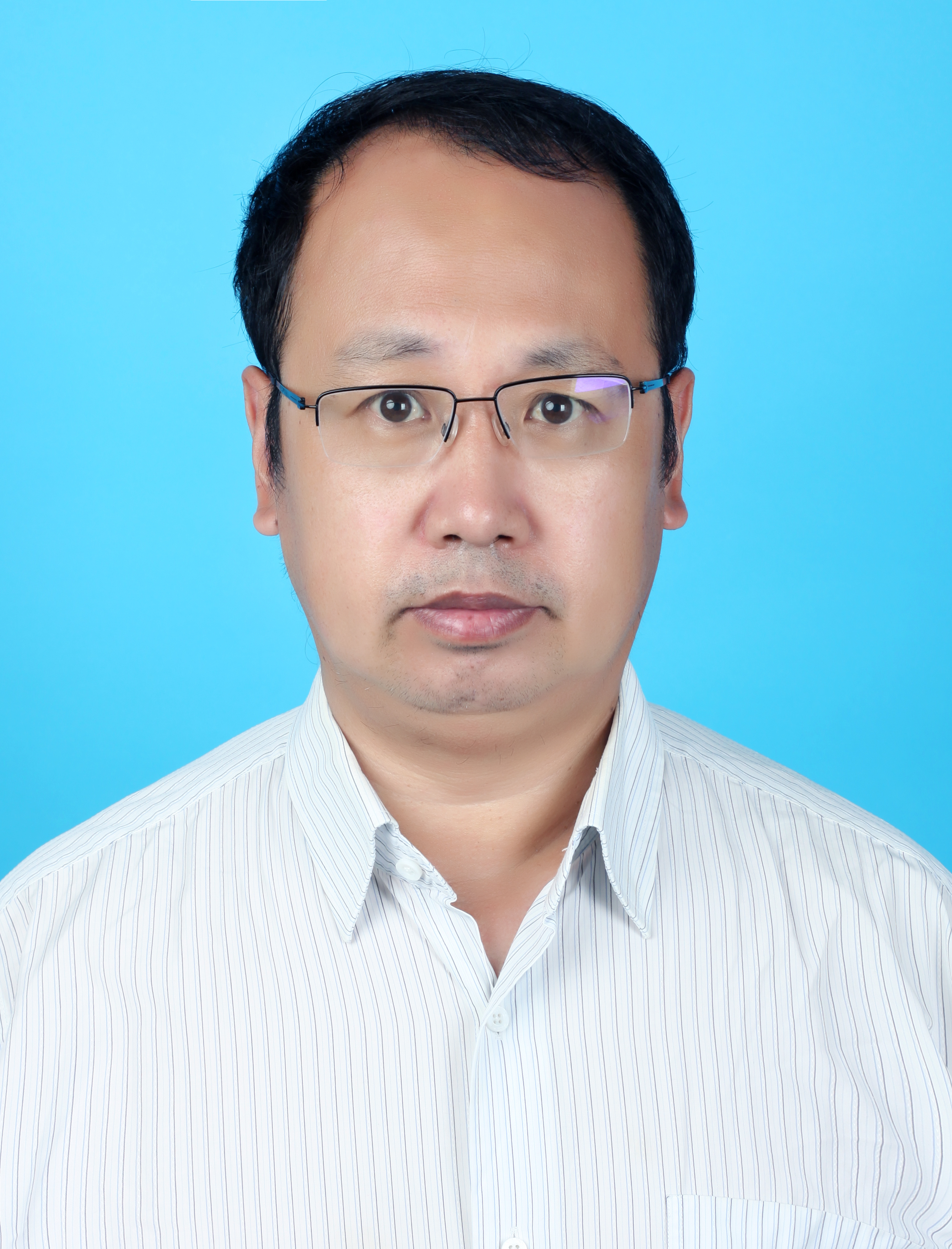}}]
	{Ke Lu} (Senior Member, IEEE) was born in Ningxia on March 13, 1971. He received the master’s and Ph.D. degrees from the Department of Mathematics and Department of Computer Science, Northwest University, Xi'an, Shanxi, China, in July 1998 and July 2003, respectively.
	He worked as a Post-Doctoral Fellow with the Institute of Automation, Chinese Academy of Sciences, Beijing, China, from July 2003 to April 2005. Currently, he is a Distinguished Professor with the University of the Chinese Academy of Sciences, Beijing. He is also a Double-hired Professor with the Pengcheng Laboratory, Shenzhen, Guangdong, China. His current research areas focus on computer vision, 3-D image reconstruction, and computer graphics.
\end{IEEEbiography}

\begin{IEEEbiography}
	[{\includegraphics[width=1in,height=1.25in,clip,keepaspectratio]{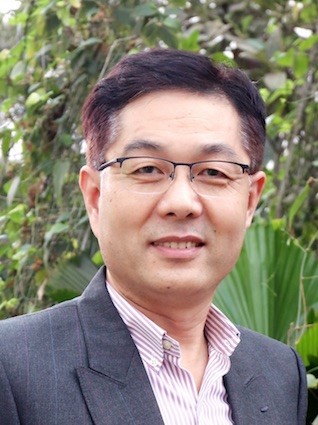}}]{Cuntai Guan}(Fellow, IEEE) received his Ph.D. degree from Southeast University, China, in 1993. He is a President’s Chair Professor in the School of Computer Science and Engineering, Nanyang Technological University, Singapore. He is the Director of the Artificial Intelligence Research Institute, Director of the Centre for Brain-Computing Research, and Co-Director of S-Lab for Advanced Intelligence. His research interests include brain-computer interfaces, machine learning, medical signal and image processing, artificial intelligence, and neural and cognitive rehabilitation. He is a recipient of the Annual BCI Research Award (first prize), King Salman Award for Disability Research, IES Prestigious Engineering Achievement Award, Achiever of the Year (Research) Award, and Finalist of President Technology Award. He is also an elected Fellow of the US National Academy of Inventors (NAI), the Academy of Engineering Singapore (SAEng), and the American Institute for Medical and Biological Engineering (AIMBE).
\end{IEEEbiography}

\end{document}